\documentclass[journal]{IEEEtran}

\usepackage[T1]{fontenc}

\usepackage{hyperref}
\usepackage{amsmath}
\usepackage[percent]{overpic}

\usepackage{algorithm}
\usepackage{algpseudocode}
\usepackage{color} 
\usepackage[cmintegrals]{newtxmath}
\usepackage{times}
\usepackage{epsfig}
\usepackage{graphicx}
\usepackage[percent]{overpic}
\usepackage{pict2e}
\usepackage{booktabs}
\usepackage{diagbox}

\usepackage{amsmath}
\usepackage{amssymb}
 \usepackage{array,multirow,graphicx}
\usepackage{amsmath,amssymb}
\usepackage{epsfig}
\graphicspath{{Figs/}}

\usepackage{array}
\newcolumntype{L}[1]{>{\raggedright\let\newline\\\arraybackslash\hspace{0pt}}m{#1}}
\newcolumntype{C}[1]{>{\centering\let\newline\\\arraybackslash\hspace{0pt}}m{#1}}
\newcolumntype{R}[1]{>{\raggedleft\let\newline\\\arraybackslash\hspace{0pt}}m{#1}}

\def\R{\mathbb R}

\newcommand{\etal}{\textit{et al.}}

\begin{document}

\title{Deep Multimodal Subspace Clustering Networks}

\author{{Mahdi~Abavisani,~\IEEEmembership{Student Member,~IEEE} and
Vishal M.		Patel,~\IEEEmembership{Senior Member,~IEEE}} 
\thanks{M.		Abavisani is with the department of Electrical and Computer Engineering at Rutgers University, 
   Piscataway, NJ USA.		email: mahdi.abavisani@rutgers.edu.		}
\thanks{Vishal M.		Patel is with the department of Electrical and Computer Engineering at Johns Hopkins University, Baltimore, MD USA.		email: vpatel36@jhu.edu.}
}

\markboth{IEEE JOURNAL OF SELECTED TOPICS IN SIGNAL PROCESSING,~Vol.~12, No.~6, December, 2018.}%
{Shell \MakeLowercase{\textit{et al.}}: Bare Demo of IEEEtran.cls for IEEE Communications Society Journals}

\maketitle

\begin{abstract}
    We present convolutional neural network (CNN) based approaches for unsupervised multimodal subspace clustering.  The proposed framework  consists of three main stages - multimodal encoder, self-expressive layer, and multimodal decoder.  The encoder takes multimodal data as input and fuses them to a latent space representation. The self-expressive layer is responsible for enforcing the self-expressiveness property and acquiring an affinity matrix corresponding to the data points.   The decoder reconstructs the original input data.  The network uses the distance between the decoder's reconstruction and the original input in its training.  We investigate early, late and intermediate fusion techniques and propose three different encoders corresponding to them for spatial fusion.  The self-expressive layers and multimodal decoders are essentially the same for different spatial fusion-based approaches.  In addition to various spatial fusion-based methods, an affinity fusion-based network is also proposed in which the self-expressive layer corresponding to different modalities is enforced to be the same.     Extensive experiments on three datasets show that the proposed methods significantly outperform the state-of-the-art multimodal subspace clustering methods.      
	
 \end{abstract}


\IEEEpeerreviewmaketitle

\section{Introduction}
\label{sec:intro}

\IEEEPARstart {M}{any} practical applications in image processing, computer vision, and speech processing require one to process very high-dimensional data.  However, these data often lie in a low-dimensional subspace.  For instance,  facial images with variation in illumination ~\cite{BasriJacobs},  handwritten digits~\cite{digitsDim} and trajectories of a rigidly moving object in a video~\cite{KanadeMotionSeg} are examples where the high-dimensional data can be represented by low-dimensional  subspaces.  Subspace clustering algorithms essentially use this fact to find clusters in different subspaces within a dataset~\cite{SC_vidal}.   In other words, in a subspace clustering task, given the data from a union of subspaces, the objective is to find the number of subspaces, their dimensions, the segmentation of the data and a basis for each subspace~\cite{SC_vidal}.  This problem has numerous applications including motion segmentation~\cite{wu2001multibody}, unsupervised image segmentation~\cite{yang2008unsupervised}, image representation and compression~\cite{hong2006multiscale} and face clustering \cite{ho2003clustering}.

\begin{figure}[t]
	\includegraphics[width=.48\textwidth]{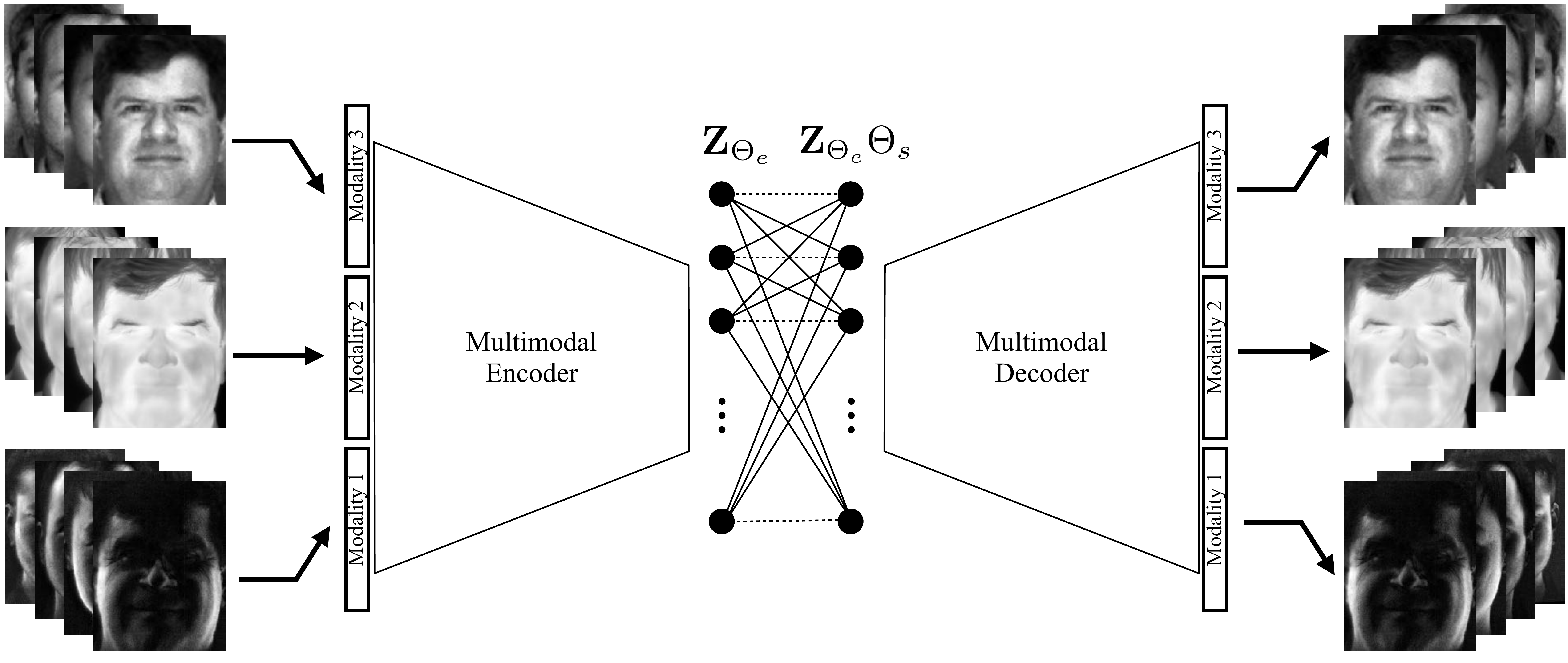}
	\caption{An overview of the proposed deep multimodal subspace clustering framework.	Note that the network consists of three blocks: a multimodal encoder, a self-expressive layer, and a multimodal decoder.	The weights in the self-expressive layer, $\boldsymbol{\Theta}_s$, are used to construct the affinity matrix. We present several models for the multimodal encoder.}	\label{fig:overvieww}
\end{figure}

Various subspace clustering methods have been proposed in the literature \cite{LLMC, LSA, SSC_CVPR, LRR, LRSC,li2015structured,you2016scalable,deepsc17nips,abavisani2018adversarial,abavisani2016domain,lu2012robust,ji2014efficient}.  In particular, methods based on sparse and low-rank representation have gained a lot of attraction in recent years \cite{SSC_PAMI, LRR_PAMI_2013,li2015structured,you2016scalable, latenst_SSC_LRR, Patel_KSSC_ICIP14, LRSSC_NIPS2013, zhang2015low}.   These methods exploit the fact that noiseless data in a union of subspaces are self-expressive, i.e. each data point can be expressed as a sparse linear combination of other data points.      The self-expressiveness property was also recently investigated in \cite{deepsc17nips} to develop a deep convolutional neural network (CNN) for subspace clustering.  This deep learning-based method was shown to significantly outperform the state-of-the-art subspace clustering methods.  

In the case where the data consists of multiple modalities or views, multimodal subspace clustering methods can be employed to simultaneously cluster the data in the individual modalities according to their subspaces \cite{MultiTask_LRR_ICCV2011,Chaudhuri,Kumar,Zhao,8052206, Convex_multiview_NIPS2012_4632, CVPR2014_multi_feature_SC, multiview_SC_AAAI_2014, desa05spectral,abavisani2018multimodal}.  Some of the multimodal subspace clustering methods make use of the kernel trick to map the data onto a high-dimensional feature space to achieve better clustering \cite{abavisani2018multimodal}.

Motivated by the recent advances in deep subspace clustering \cite{deepsc17nips} as well as multimodal data processing using CNNs \cite{ngiam2011multimodal,srivastava2012multimodal,ramachandram2017deep, kahou2016emonets,jain2014modeep,valada2016deep,antol2015vqa,donahue2015long,perera2017in2i,di2017large}, in this paper, we propose a different approach to the problem of multimodal subspace clustering.   We present a novel CNN-based autoencoder approach in which a fully-connected layer is introduced between the encoder and the decoder which mimics the self-expressiveness property that has been widely used in various subspace clustering algorithms. 

Figure~\ref{fig:overvieww} gives an overview of the proposed deep multimodal subspace clustering framework.  The self-expressive layer is responsible for enforcing the self-expressiveness property and acquiring an affinity matrix corresponding to the data points.   The decoder reconstructs the original input data from the latent features.  The network uses the distance between the decoder's reconstruction and the original input in its training.

For encoding the multimodal data into a latent space, we investigate three different spatial fusion techniques based on late, early and intermediate fusion.  These fusion techniques are motivated by the deep multimodal learning methods in supervised learning tasks \cite{kiela2018efficient, feichtenhofer2016convolutional}, that provide the representation of modalities across spatial positions.  In addition to the spatial fusion methods, we propose an affinity fusion-based network in which the self-expressive layer corresponding to different modalities is enforced to be the same.  For both spatial and the affinity fusion-based methods, we formulate an end-to-end training objective loss.  

Key contributions of our work are as follows:
\begin{itemize}
	\item Deep learning-based multimodal subspace clustering framework is proposed in which the self-expressiveness property is encoded in the latent space by using a fully connected layer.  
	\item Novel encoder network architectures corresponding to late, early and intermediate fusion are proposed for fusing multimodal data. 
	\item An affinity fusion-based network architecture is proposed in which the self-expressive layer is enforced to have the same weights across latent representations of all the modalities.   
	\end{itemize}
To the best of our knowledge, this is the first attempt that proposes to use deep learning for multimodal subspace clustering. Furthermore, the proposed method obtains the state-of-the-art results on various  multimodal subspace clustering datasets.  Code is available at: \url{https://github.com/mahdiabavisani/Deep-multimodal-subspace-clustering-networks}.

This paper is organized as follows. Related works on subspace clustering and multimodal learning are presented in Section~\ref{sec:background}.  The proposed spatial fusion-based and affinity fusion-based multimodal subspace clustering methods are presented in Section~\ref{sec:method} and \ref{sec:affinity}, respectively. Experimental results are presented in Section~\ref{sec:exper}, and finally, Section~\ref{sec:con} concludes the paper with a brief summary.

\section{Related Work}\label{sec:background}
In this section, we review some related works on subspace clustering and multimodal learning.  

\subsection {Sparse and Low-rank Representation-based Subspace Clustering} 
	  Let $\mathbf{X}=[\mathbf{x}_{1},\cdots,\mathbf{x}_{N}]\in \R^{D\times N}$ be a collection of $N$ signals $\{\mathbf{x}_{i}\in \R^{D}\}_{i=1}^{N}$ drawn from a union of $n$ linear subspaces $\mathcal{S}_{1} \cup \mathcal{S}_{2} \cup \cdots \cup \mathcal{S}_{n}$ of dimensions $\{d_{\ell}\}_{\ell=1}^{n}$ in $\R^{D}$.
  Given $\mathbf{X}$, the task of subspace clustering is to find sub-matrices $\mathbf{X}_{\ell}\in \R^{D\times N_{\ell}}$ that lie in $\mathcal{S}_{\ell}$ with $N_{1}+N_{2}+\cdots+N_{n}=N.$
    The sparse subspace clustering (SSC)~\cite{SSC_PAMI} and low-rank representations-based subspace clustering (LRR)~\cite{LRR_PAMI_2013} algorithms exploit the fact that noiseless data in a union of subspaces are \emph{self-expressive}.	In other words, it is assumed that each data point can be represented as a linear combination of other data points. Hence, these algorithms aim to find the sparse or low-rank matrix $\mathbf{C}$ by solving the following optimization problem	
\begin{equation}\label{eq:SSC1}
 \min_{\mathbf{C}}\|\mathbf{C}\|_{p}+\frac{\lambda}{2}\|\mathbf{X}-\mathbf{X}\mathbf{C}\|_{F}^{2},
\end{equation}
where $\|.\|_{p}$ is the $\ell_{1}$-norm in the case of SSC \cite{SSC_PAMI} and the nuclear norm in the case of LRR \cite{LRR_PAMI_2013}.	Here, $\lambda$ is a regularization parameter. 	In addition, to prevent the trivial solution $\mathbf C = \mathbf I$, an additional constraint of diag$(\mathbf{C})=\mathbf{0}$ is added to the above optimization problem in the case of SSC.	  Once $\mathbf{C}$ is found, spectral clustering methods \cite{NG_spectralCluster} are applied on the affinity matrix  $\mathbf{W}=|\mathbf{C}|+|\mathbf{C}|^T$ to obtain the segmentation of the data $\mathbf{X}$.	    

Non-linear versions of the SSC and LRR algorithms have also been proposed in the literature \cite{latenst_SSC_LRR, Patel_KSSC_ICIP14}.

\subsection{Deep Subspace Clustering} 
\begin{figure}[t]
\centering	\begin{overpic}[width=.48\textwidth,tics=5]{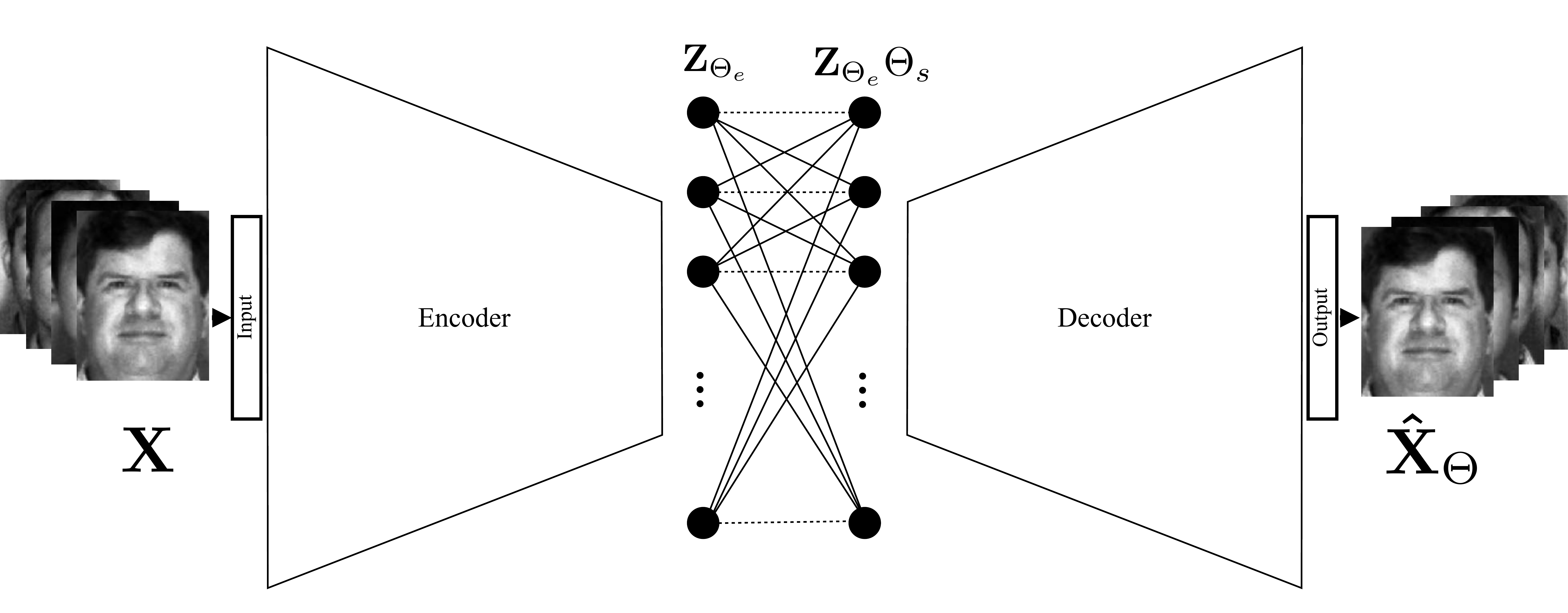}
\put (92.4,10.3) {\tiny{$\sim$}}
\end{overpic}
	\caption{An overview of the DSC framework proposed in \cite{deepsc17nips} for unimodal subspace clustering.	}
	\label{fig:dmscn}
\end{figure}

The deep subspace clustering network (DSC)~\cite{deepsc17nips} explores the self-expressiveness property by embedding the data into a latent space using an encoder-decoder type network.  Figure~\ref{fig:dmscn} gives an overview of the DSC method for unimodal subsapce clustering.  The method optimizes an objective similar to that of \eqref{eq:SSC1} but the matrix $\mathbf{C}$ is approximated using a trainable dense layer embedded within the network.  Let us denote the parameters of the self-expressive layer as $\boldsymbol{\Theta}_{s}$.  Note that these parameters are essentially the elements of $\mathbf{C}$ in \eqref{eq:SSC1}.  The following loss function is used to train the network
\begin{align}\label{eq:DSC}
\nonumber \min_{\tilde{\boldsymbol{\Theta}}} \;  \|\boldsymbol{\Theta}_s\|_{p}+\frac{\lambda_1}{2}\|\mathbf{Z}_{\boldsymbol{\Theta_e}}-\mathbf{Z}_{\boldsymbol{\Theta_e}}\boldsymbol{\Theta}_s\|_{F}^{2} &+  \frac{\lambda_2}{2} \| \mathbf{X} - \mathbf{\hat X}_{\tilde{\boldsymbol{\Theta}}}\|,\\
&\;\; \text{s.t.  } \text{diag}(\boldsymbol{\Theta}_{s})=\mathbf{0}, 
\end{align}
where $\mathbf{Z}_{\boldsymbol{\Theta}_e}$ denotes the output of the encoder, and $\hat{\mathbf{X}}_{\tilde{\boldsymbol{\Theta}}}$ is the reconstructed signal at the output of the decoder.  Here, the network parameters 	$\tilde{\boldsymbol{\Theta}}$ consist of  encoder parameters $\boldsymbol{\Theta}_{e}$, decoder parameters $\boldsymbol{\Theta}_{d}$ and self-expressive layer parameters $\boldsymbol{\Theta}_{s}$. Here, $\lambda_{1}$ and $\lambda_{2}$ are two regularization parameters.

\begin{figure*}[t]
\centering   \begin{overpic}[width=.8\textwidth,tics=5]{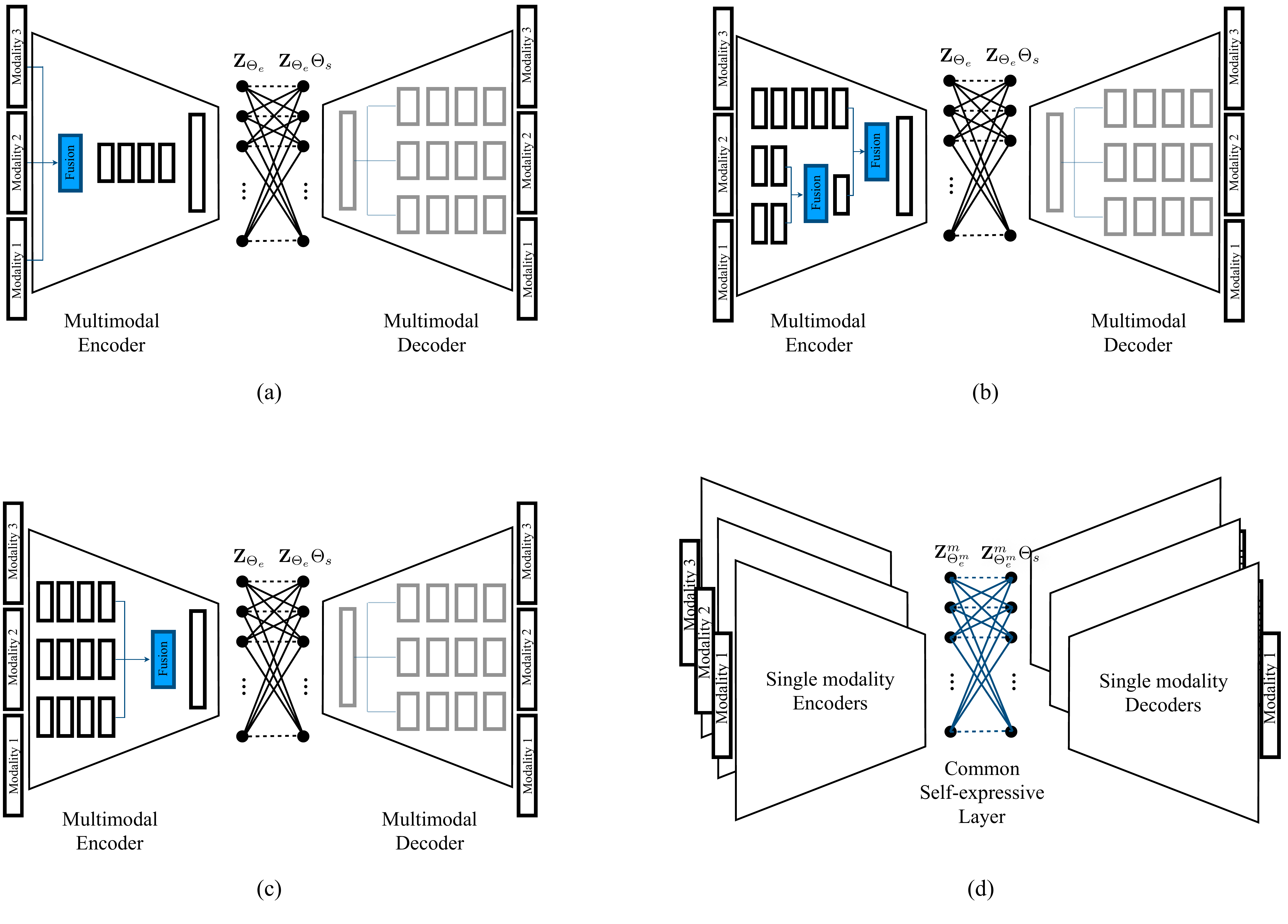}
\end{overpic}
\vskip -20pt\caption{Different network architectures corresponding to (a) early fusion, (b) intermediate fusion, and (c) late fusion.  Note that in all the spatial fusion-based networks (a)-(c), the overall structure for the self-expressive layer and the multimodal decoder remain the same.	(d) Network architecture corresponding to affinity fusion. In this case, the encoder and decoder are trained separately for each modality, but are forced to have the same self-expressive layer.}
\label{fig:spatialfusion}
\end{figure*}

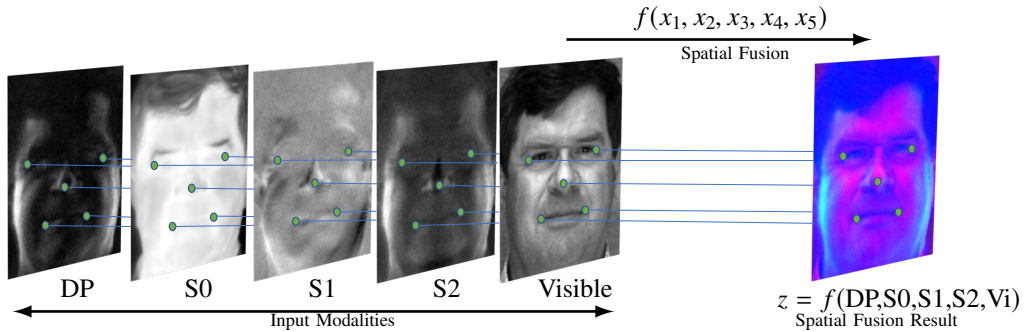
\begin{figure*}[t]
\centering \begin{overpic}[width=0.7\textwidth,tics=5]{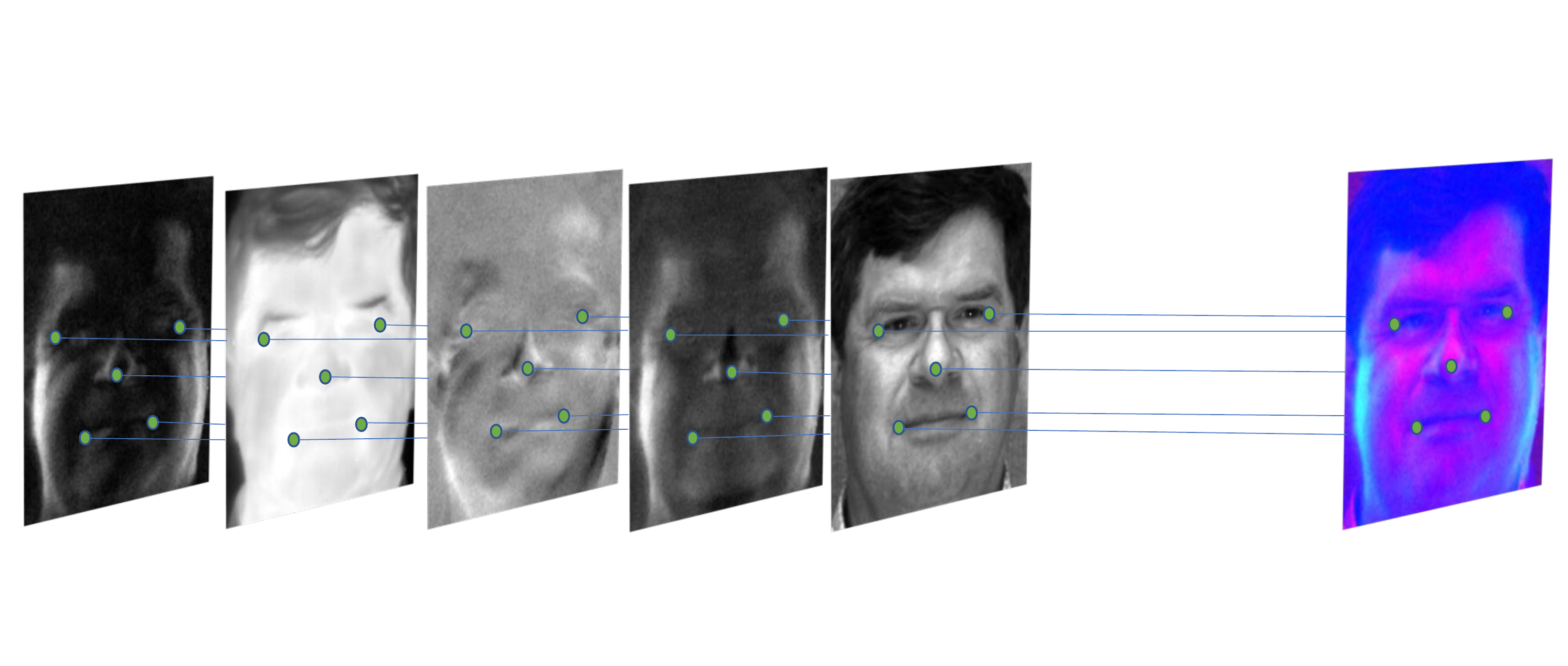}
\put (7,7) {{DP}}
\put (20,7) {{S0}}
\put (33,7) {{S1}}
\put (46,7) {{S2}}
\put (57,7) {{Visible}}
\put (82,6) {$z=f($DP,S0,S1,S2,Vi$)$}
\put (84,4) {\scriptsize{Spatial Fusion Result}}
\put (29,4) {\scriptsize{Input Modalities}}
\put (72,32) {\scriptsize{Spatial Fusion}}
\put (67,35.5) {$f(x_1,x_2,x_3,x_4,x_5)$}
\linethickness{1pt}
\put(60,34){\vector(1,0){32}}
\put(3,5.5){\vector(1,0){62}}
\put(3,5.5){\vector(-1,0){1}}
\end{overpic}
\vskip -20pt\caption{In spatial fusion methods each location of the fusion is related to the input values at the same location.		In this especial case, the facial components (i.e. eyes, nose and mouth) are aligned across all the modalities (i.e. DP, S0, S1, S2, Visible).}
\label{fig:spatial_fusion}
\end{figure*}

\subsection{Multimodal Subspace Clustering}
A number of multimodal and multiview subspace clustering approaches have been developed in recent years.  Bickel~\etal~  introduced an Expectation Maximization (EM) and agglomerative multiview clustering methods in \cite{CVPR2014_multi_feature_SC}.  White~\etal~\cite{Convex_multiview_NIPS2012_4632} provided a convex reformulation of multiview subspace learning that as opposed to local formulations enables global learning.  Some algorithms use dimensionality reduction methods such as Canonical Correlation Analysis (CCA) to project the multiview data onto a low-dimensional subspace for clustering \cite{Chaudhuri, multiview_SC_AAAI_2014}.  Some other multimodal methods are specifically designed for two views and can not be easily generalized to multiple views \cite{Bickel, desa05spectral}. Kumar \etal~\cite{Kumar} proposed a co-regularization method that enforces the clusterings to be aligned in different views.  Zhao \etal~\cite{Zhao} use output of clustering in one view to learn discriminant subspaces in another view.   A multiview subspace clustering method, called Low-rank Tensor constrained Multiview Subspace Clustering (LT-MSC) was recently proposed in~\cite{zhang2015low}.		 In the LT-MSC method, all the subspace representations are integrated into a low-rank tensor, which captures the high order correlations underlying multiview data.		  In~\cite{diversity_multiview_SC}, a diversity-induced multiview subspace clustering was proposed in which the Hilbert Schmidt independence criterion was utilized to explore the complementarity of multiview representations.		  Recently,~\cite{cao2015constrained} proposed a constrained multi-view video face clustering (CMVFC) framework in which pairwise constraints are employed in both sparse subspace representation and spectral clustering procedures for multimodal face clustering.		  A collaborative image segmentation framework,  called  Multi-task  Low-rank  Affinity  Pursuit (MLAP) was proposed in~\cite{MultiTask_LRR_ICCV2011}.		 In this method, the sparsity-consistent low-rank affinities from the joint decompositions
of multiple feature matrices into pairs of sparse and low-rank matrices are exploited for segmentation.		

\subsection{Deep Multimodal Learning} In multimodal learning problems, the idea is to use the complementary information provided by the different modalities to enhance the recognition performance.Supervised deep multimodal learning was first introduced in \cite{ngiam2011multimodal},  \cite{srivastava2012multimodal}, and  has gained a lot of attention in recent years~\cite{kim2016multimodal,neverova2016moddrop,kahou2016emonets}.

Keila \etal~\cite{kiela2018efficient} investigated deep multimodal classification of large-scaled datasets.   They,  compared a number of multimodal fusion methods in terms of accuracy and computational efficiency, and provided analysis regarding the  interpretability of multimodal classification models. Feichtenhofer \etal~\cite{feichtenhofer2016convolutional} proposed a convolutional fusion method for two stream 3D networks.  They explored multiple fusion functions within deep architectures and studied the importance of learning the correspondences between spatial and temporal feature maps.  Various deep supervised multimodal fusion approaches have also been proposed in the literature for different applications including medical image analysis applications~\cite{simonovsky2016deep,liu2015multimodal}   visual recognition~\cite{jain2014modeep,kahou2016emonets} and   visual question answering~\cite{kim2016multimodal,antol2015vqa}. We refer readers to \cite{ramachandram2017deep} for more detailed survey of various  deep supervised multimodal fusion methods.


While most of the deep multimodal approaches have reported improvements in the supervised tasks, to the best of our knowledge, there is no deep multimodal learning method specifically designed for unsupervised subspace clustering.

\section{Spatial Fusion-based Deep Multimodal Subspace Clustering}\label{sec:method}
In this section, we present details of the proposed spatial fusion-based networks for unsupervised subspace clustering.  Spatial fusion methods find a joint representation that contains complementary information from different modalities.  The joint representation has a spatial correspondence to every modality.  Figure~\ref{fig:spatial_fusion} shows a visual example of spatial fusion where five different modalities (DP, S0, S1, S2, Visible) are combined to produce a fused result $z$.  The spatial fusion methods are especially popular in supervised multimodal learning applications~\cite{kiela2018efficient, feichtenhofer2016convolutional}. We investigate applying these fusion techniques to our problem of deep subspace clustering.		

An essential component of such methods is the fusion function that merges the information from multiple input representations and returns a fused output. In the case of deep networks,  flexibility in the choice of fusion network leads to different models.  In what follows, we investigate several network designs and spatial fusion functions for multimodal subspace clustering. Then, we formulate an end-to-end training objective for the proposed networks.

\subsection{Fusion Structures}
We build our deep multimodal subspace clustering networks based on the architecture proposed in \cite{deepsc17nips} for unimodal subspace clustering.  Our framework consists of three main components: an encoder, a fully connected self-expressive layer, and a decoder.		   We propose to achieve the spatial fusion using an encoder and the fused representation is then fed to a self-expressive layer which essentially exploits the self-expressiveness property of the joint representation.		  The joint representation resulting from the output of the self-expressive layer is then fed to a multimodal decoder that reconstructs the different modalities from the joint latent representation.		

For the case of $M$ input modalities, the decoder consists of $M$ branches, each reconstructing one of the modalities.  The encoders on the other hand, can be designed such that they achieve  early, late or intermediate fusion.  Early fusion refers to the integration of multimodal data in the stage of feature level before feeding them to the network.		  Late fusion, on the other hand, involves the integration of multimodal data in the last stage of the network. The flexibility of deep networks also offers the third type of fusion known as the intermediate fusion, where the feature maps from the intermediate layers of a network are combined to achieve better joint representation. 	Figures \ref{fig:spatialfusion} (a), (b) and (c) give an overview of deep multimodal subspace clustering networks with different spatial fusion structures.		 Note that the multimodal decoder's structure remains the same in all three cases.  It is worth mentioning that in the case of intermediate fusion, it is a common practice to aggregate the weak or correlated modalities at earlier stages and combine the remaining strong modalities at the in-depth stages~\cite{ramachandram2017deep}.

\subsection{Fusion Functions}
Assume for a particular data point, $x_i$, there are $M$ feature maps corresponding to the representation of different modalities.		A fusion function $f : \{x^1,x^2.		\cdots, x^M\} \rightarrow z$ fuses the $M$ feature maps and produces an output $z$.		   For simplicity we assume that all the input feature maps have the same dimension of $\R^{H \times W \times d^{in}}$, and the output has the dimension of 
$\R^{H \times W \times d^{out}}$.		  In fact, deep network structures offer the design option for having feature maps with the same dimensions.		 We use  $z_{i,j,k}$ and $x^{m}_{i,j,k}$  to denote the value in the spatial position $(i,j,k)$ in the output and the $m$th input feature map, respectively.  Various fusion functions can be used to combine the input feature maps.  Below, we investigate a few.

\subsubsection{Sum fusion $z=\text{sum}(x^1,x^2.		\cdots, x^M)$} computes the sum of the feature maps at the same special positions as follows  
\begin{equation}\label{eq:sum}
z_{i,j,k}=\sum_{m=1}^{M}{x^{m}_{i,j,k}}.
\end{equation}

\subsubsection{Maxpooling function $z=\text{max}(x^1,x^2.		\cdots, x^M)$}  returns the maximum value of the corresponding location in the input feature maps as follows
\begin{equation}\label{eq:sum}
z_{i,j,k}=\text{Max}\{x^1_{i,j,k},x^2_{i,j,k}.		\cdots, x^M_{i,j,k}\}.
\end{equation}

\subsubsection{Concatenation function $z=\text{cat}(x^1,x^2.		\cdots, x^M)$} constructs the output by concatenating the input feature maps as follows		
\begin{equation}\label{eq:sum}
z=[x^1,x^2.		\cdots, x^M],
\end{equation}
where each input has the dimension $\R^{H \times W \times d_{in}}$ and the output has the dimension $\R^{H \times W \times (d_{in}\times M)}$.  Note that these fusion functions are denoted as ``Fusion" in blue boxes in Figure~\ref{fig:spatialfusion} (a)-(c).



\subsection{End-to-End Training Objective}
Given $N$ paired data samples $\{\mathbf{x}_{i}^{1}, \mathbf{x}_{i}^{2}, \cdots, \mathbf{x}_{i}^{M}\}_{i=1}^{N}$ from $M$ different modalities, define the corresponding data matrices as $\mathbf{X}^{m}=[\mathbf{x}_{1}^{m}, \mathbf{x}_{s}^{m}, \cdots, \mathbf{x}_{N}^{m}],  \;\;m \in \{1,\cdots,M\}$.  Regardless of the network structure and the fusion function of choice, let  $\boldsymbol{\Theta}_{M.e}$ denote the parameters of the multimodal encoder.		 Similarly, let $\boldsymbol{\Theta}_{s}$ be the self-expressive layer parameters and $\boldsymbol{\Theta}_{M.d}$ be the multimodal decoder parameters.  Then the proposed spatial fusion models can be  trained end-to-end using the following loss function 
\begin{align}\label{eq:SFDMSC}
\nonumber \min_{\boldsymbol{\Theta}}  \|\boldsymbol{\Theta}_{s}\|_{p}+\frac{\lambda_1}{2}\|\mathbf{Z}_{\boldsymbol{\Theta}_{M.e}} - \mathbf{Z}_{\boldsymbol{\Theta}_{M.e}}\boldsymbol{\Theta}_{s}\|_{F}^{2} &+  \frac{\lambda_2}{2} \sum^{M}_{m=1} \| \mathbf{X}^m - \hat{\mathbf{X}}^m_{\boldsymbol{\Theta}}\|\\
&\text{s.t  }\text{diag}(\boldsymbol{\Theta}_{s})=\mathbf{0}, 
\end{align}
where $\boldsymbol{\Theta}$ denotes all the training network parameters including $\boldsymbol{\Theta}_{M.e}$, $\boldsymbol{\Theta}_{s}$  and $\boldsymbol{\Theta}_{M.d}$.		The joint representation is denoted by $\mathbf{Z}_{\boldsymbol{\Theta}_{M.e}}$, and $\hat{\mathbf{X}}^m_{\boldsymbol{\Theta}}$ is the reconstruction of $\mathbf{X}^m$. Here, $\lambda_{1}$ and $\lambda_{2}$ are two regularization parameters,  and  $\|\cdot \|_{p}$  can be either $\ell_1$ or $\ell_2$ norm.

\section{Affinity Fusion-based Deep Multimodal Subspace Clustering}\label{sec:affinity}

\begin{figure*}[t]
\centering \begin{overpic}[width=0.85\textwidth,tics=3]{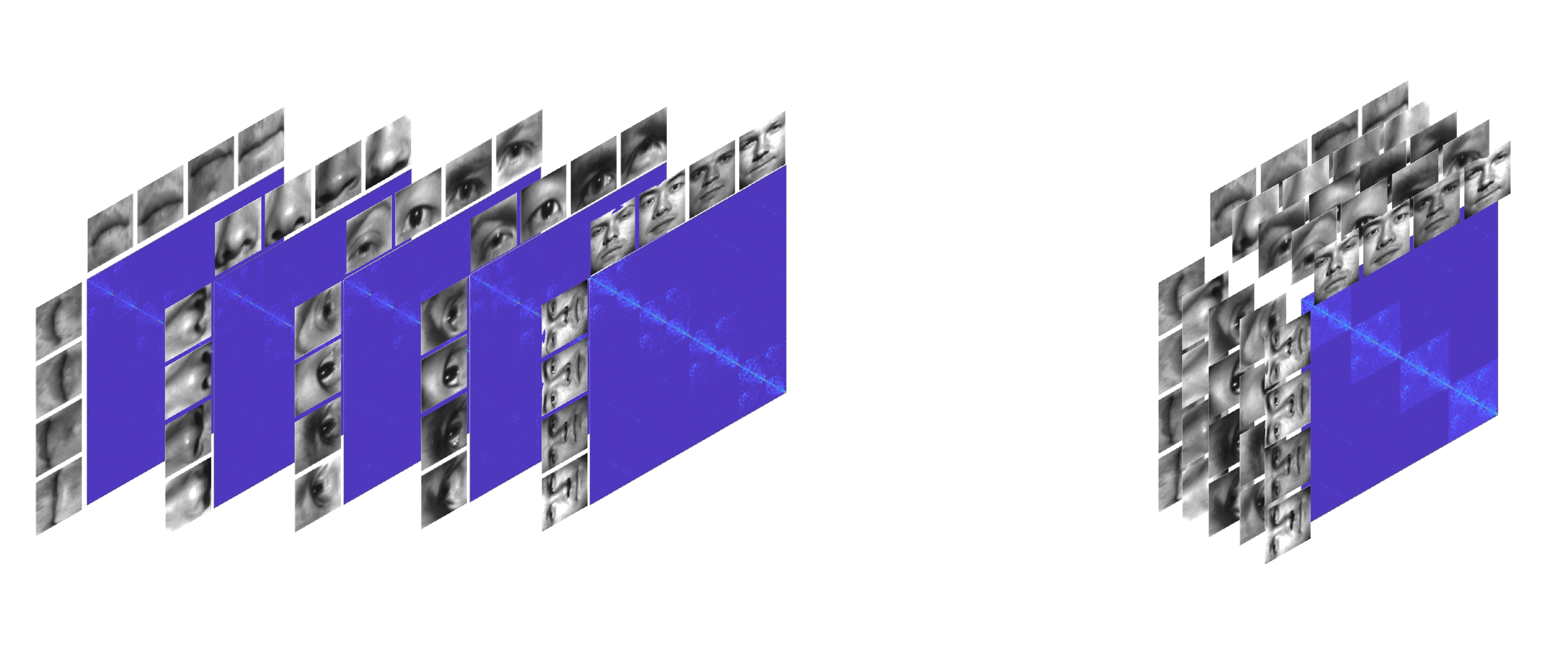}
\put (5,6) {\rotatebox{35}{\scriptsize{Mouths}}}
\put (13,6) {\rotatebox{35}{\scriptsize{Noses}}}
\put (21,6) {\rotatebox{35}{\scriptsize{L-eyes}}}
\put (30,6) {\rotatebox{35}{\scriptsize{R-eyes}}}
\put (39,6) {\rotatebox{35}{\scriptsize{Faces}}}
\put (79,4) {\scriptsize{Shared Affinity}}
\put (12,3) {\scriptsize{Individual Affinities of the Input Modalities}}
\put (58,23) {\scriptsize{Affinity Fusion}}

\linethickness{1pt}
\multiput(55,22)(2,0){6}{\line(1,0){1}}
\put(67,22){\vector(1,0){3}}
\put(15,4.5){\vector(1,0){36}}
\put(15,4.5){\vector(-1,0){11}}
\end{overpic}
\vskip -15pt\caption{An example of affinity fusion.	 Affinities corresponding to different modalities are combined to have only a single shared affinity. This method does not relay on spatial relation across different modalities.		 Instead, it aggregates the similarities among data points across different modalities and returns a shared affinity.}
\label{fig:affinity_fusion}
\end{figure*}
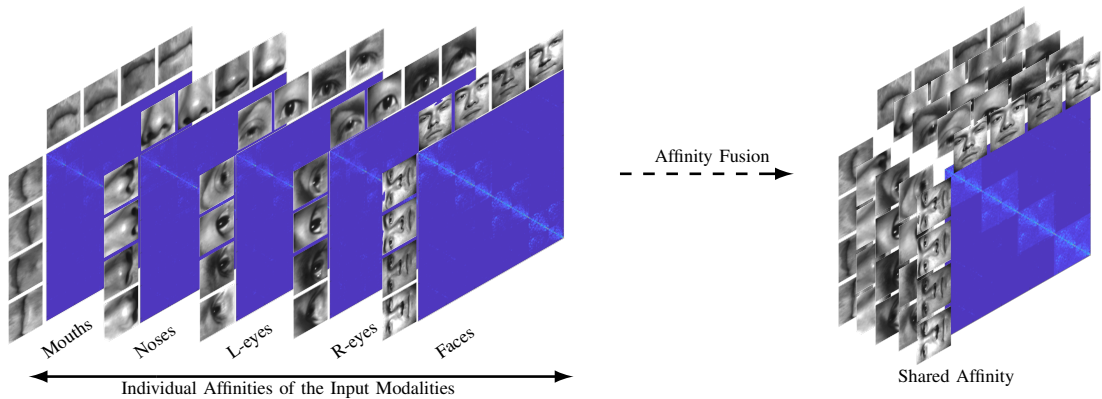


In this section, we propose a new method for fusing the affinities across the data modalities to achieve better clustering.  Spatial fusion methods require the samples from different modalities to be aligned (see Figure~\ref{fig:spatial_fusion}) to achieve better clustering.  In contrast, the proposed affinity fusion approach combines the similarities from the self-expressive layer to obtain a joint representation of the multimodal data.  This is done by enforcing the network to have a joint affinity matrix.  This avoids the issue of having aligned data or increasing the dimensionality of the fused output (i.e. concatenation).   The motivation for enforcing a shared affinity matrix is that similar~(dissimilar) data in one modality should be similar~(dissimilar) in the other modalities as well.		  Figure~\ref{fig:affinity_fusion} shows an example of the proposed affinity fusion method by forcing the modalities to share the same affinity matrix.

In the DSC framework \cite{deepsc17nips}, the affinity matrix is calculated from the self-expressive layer weights as follows $$\mathbf{W}=|\boldsymbol{\Theta}_s^T| + |\boldsymbol{\Theta}_s^T|,$$ where $\boldsymbol{\Theta}_s$ corresponds to the self-expressive layer weights learned by an end-to-end training strategy \cite{deepsc17nips}.		Thus a shared $\boldsymbol{\Theta}_s$ results in a common $\mathbf{W}$ across the modalities. We enforce the modalities to share a common $\boldsymbol{\Theta}_s$ while having different encoders,  decoders and the latent representations. 

\subsection{Network Structure}
For an $M$ modality problem, we propose to stack $M$ parallel DSC networks, where they share a common self-expressive layer.		 In this model, per each modality one encoder-decoder network is trained.		In contrast to the spatial fusion models that only have one joint latent representation,  this model results in $M$ distinct latent representations corresponding to $M$ different modalities.		 The latent representations are connected together by sharing the self-expressive layer.		 The optimal self-expressive layer should be able to jointly exploit the self-expressiveness property across all the $M$ modalities.		   Figure \ref{fig:spatialfusion} (d) gives an overview of the proposed affinity fusion-based network architecture.

\subsection{End-to-End Training}
 We propose to find the shared self-expressive layer weights by training the network with the following loss 
\begin{align}\label{eq:DMSC}
\nonumber \min_{\boldsymbol{\Theta}}\;\;  \|\boldsymbol{\Theta}_{s}\|_{p}&+\frac{\lambda_1}{2}\sum_{m=1}^{M}\|\mathbf{Z}^{m}_{\boldsymbol{\Theta}^m_e}- \mathbf{Z}^m_{\boldsymbol{\Theta}^m_e}\boldsymbol{\Theta}_{s}\|_{F}^{2}\\
&+  \frac{\lambda_2}{2}\sum_{m=1}^{M} \| \mathbf{X}^m - \hat{ \mathbf{X}}^{m}_{\boldsymbol{\Theta}^m}\|
\text{  s.t. } \text{diag}(\boldsymbol{\Theta}_s)=\mathbf{0}, 
\end{align}
where $\boldsymbol{\Theta}_{s}$ is the common self-expressive layer weighs.	Here, $\lambda_{1}$ and $\lambda_{2}$ are regularization parameters.  $\mathbf{Z}^{m}_{\boldsymbol{\Theta}^m_e}$ and $\hat{\mathbf{X}}^m_{\boldsymbol{\Theta}^m}$  are respectively the latent space representation and the reconstructed decoder's output corresponding to $\mathbf{X}^m$.		    $\boldsymbol{\Theta}^m$ denotes the network parameters corresponding to the $m$th modality and $\boldsymbol{\Theta}$ indicates to all the trainable parameters.  Minimizing \eqref{eq:DMSC} encourages the networks to learn the latent representations that share the same affinity matrix.	

Algorithm~\ref{alg:DMSC} summarizes the proposed spatial fusion and affinity fusion-based subspace clustering methods.	Details of different network architectures used in this paper are given in Appendix \ref{sec:appendix}. 

\begin{algorithm}[t]
\caption{Spatial and affinity fusion algorithms}\label{alg:DMSC}
\begin{algorithmic}[1]
\Procedure{{DMSC}}{$\{\mathbf{X}^m\}^M_{m=1}$, $\lambda_{1}, \lambda_{2}, \texttt{`mode'}$}
\If{\texttt{mode} = \text{Spatial fusion}} 
\State Train the networks using the loss \eqref{eq:SFDMSC}.
\ElsIf{\texttt{mode} = \text{Affinity fusion}}
\State Train the networks using the loss \eqref{eq:DMSC}.
\EndIf
\State Extract $\boldsymbol{\Theta}_{s}$ from the trained networks.
 \State Normalize the columns of $\boldsymbol{\Theta}_{s}$ as $\boldsymbol{\theta}_{si}\leftarrow \frac{\boldsymbol{\theta}_{si}}{\|\boldsymbol{\theta}_{si}\|_{\infty}}$.
\State Form a similarity graph with $N$ nodes and set the weights on the edges by $\mathbf{W}=|\boldsymbol{\Theta}_{s}|+|\boldsymbol{\Theta}_{s}^{T}|$.
\State Apply spectral clustering to the similarity graph.
\EndProcedure
\State \textbf{Output:} Segmented multimodal data.
 \end{algorithmic}
\end{algorithm}

\section{Experimental Results}
\label{sec:exper}
We evaluate the proposed deep multimodal subspace clustering methods on several real-world multimodal datasets. The following datasets are used in our experiments.  
\begin{itemize}
	\item Multiview digit clustering using the  MNIST~\cite{lecun2010mnist} and the USPS~\cite{hull1994database} handwritten digits datasets.  Here, we view images from the individual datasets as two views of the same digit.	 These datasets are considered to be spatially related but not aligned. Since the number of parameters in the self-expressive layer of a deep subspace clustering network scales quadratically with the size of the data, we randomly select 200 samples per digit to keep the networks to a tractable size. 
	\item Heterogeneous face clustering using the ARL Polarimetric face dataset~\cite{hu2016polarimetric}.  The ARL dataset contains five spatially well-aligned modalities (Visible, DP, S0, S1, S2).  
	\item Face clustering based on the facial regions using the  Extended Yale-B dataset~\cite{9ptsLight}.  We extract facial components (i.e. eyes, nose, mouth) from the images and view them as soft biometrics and use them along with the entire face for clustering.  Here, the modalities do not share any direct  spatial correspondence.
	\end{itemize}
		   
\begin{figure*}[t]
\centering \begin{overpic}[width=0.95\textwidth,tics=3]{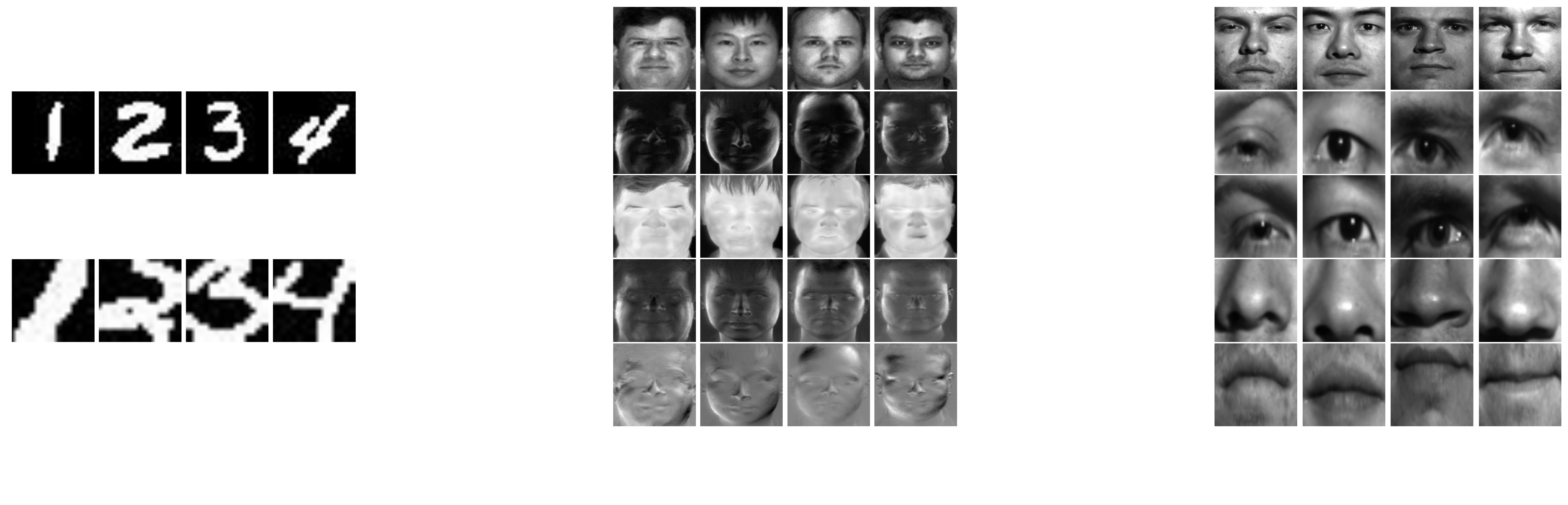}
\put (11,3) {{\scriptsize{(a)}}}
\put (49.25,3) {{\scriptsize{(b)}}}
\put (88,3) {{\scriptsize{(c)}}}
\put (-1,10) {\rotatebox{90}{\scriptsize{USPS~\cite{hull1994database}}}}
\put (-1,20.5) {\rotatebox{90}{\scriptsize{MNIST~\cite{lecun2010mnist}}}}
\put (37.5,7) {\rotatebox{90}{\scriptsize{S2}}}
\put (37.5,12) {\rotatebox{90}{\scriptsize{S1}}}
\put (37.5,17.5) {\rotatebox{90}{\scriptsize{S0}}}
\put (37.5,23) {\rotatebox{90}{\scriptsize{DP}}}
\put (37.5,27) {\rotatebox{90}{\scriptsize{Visible}}}
\put (76,6) {\rotatebox{90}{\scriptsize{Mouth}}}
\put (76,11.5) {\rotatebox{90}{\scriptsize{Nose}}}
\put (76,17) {\rotatebox{90}{\scriptsize{R-eye}}}
\put (76,22) {\rotatebox{90}{\scriptsize{L-eye}}}
\put (76,28) {\rotatebox{90}{\scriptsize{Face}}}
\end{overpic}
\vskip -20pt\caption{Sample images from (a)  MNIST~\cite{lecun2010mnist}, and USPS~\cite{hull1994database} digits datasets, (b) ARL polarimetric face dataset \cite{hu2016polarimetric}, and (c) Faces and facial components from the Extended Yale-B dataset~\cite{9ptsLight}. In our experiments, samples from all the modalities are resized to $32\times32$, and rescaled to have pixel values between 0 and 255.}
\label{fig:datasets}
\end{figure*}
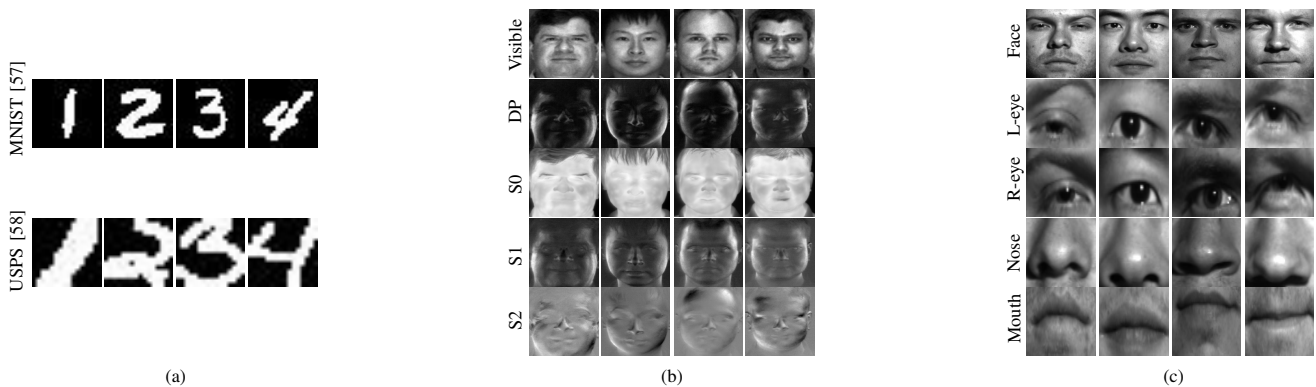

		 Figure~\ref{fig:datasets} (a), (b), and (c) show sample images from the digits, ARL and Extended Yale-B datasets, respectively. Table~\ref{tbl:datasets} gives an overview of their details. Note that as opposed to supervised methods, we do not split datasets into training and testing sets for subspace clustering. Similar to \cite{deepsc17nips}, the parameters of the deep subspace clustering networks are trained using the entire dataset. 

\begin{table}[t]
\begin{center}
\resizebox{.75\linewidth}{!}{%
\begin{tabular}{|L{1.7cm}|L{1.5cm}|C{1.1cm}|C{1.5cm}|}

\hline
  Experiment &	Dataset	& 	$\#$ of  \newline modalities	&	$\#$ of   samples \newline per modality\\
\hline
Digits &	MNIST~\cite{lecun2010mnist}, USPS~\cite{hull1994database}	& 	2	&	2000 \\
\hline
Heterogeneous  Faces &	ARL~\cite{hu2016polarimetric}	& 	5	&	2160\\
\hline
Facial \newline components &	Extended Yale-B~\cite{9ptsLight}	& 	5	&	2432 \\
\hline
\end{tabular}
}
\caption{Details of the multimodal datasets that are used in the experiments. Note that as opposed to supervised methods, we do not split datasets to training and testing sets in a deep subspace clustering task. } \label{tbl:datasets}
\end{center}
\end{table}

To investigate ability and limitations of different versions of the proposed fusion methods,  we evaluate the affinity fusion method along with a wide range of plausible spatial fusion methods based on different structure designs and fusion functions.		    For the early fusion structure, we consider the concatenation fusion function\footnote{Note that applying max-pooling and additive functions in pixel level features might result in information loss.}. 		 As for the intermediate and late fusion structures, we consider all the three presented fusion functions which results in six distinct models. Table~\ref{tbl:spatialmethods} presents the structural variations we have used for the presented spatial fusion methods and the name we assign to them when reporting their performances.	   Besides,  we compare our methods against the following state-of-the-art multimodal subspace clustering baselines: CMVFC \cite{cao2015constrained}, TM-MSC \cite{zhang2015low},  MSSC \cite{abavisani2018multimodal}, MLRR \cite{abavisani2018multimodal}, KMSSC \cite{abavisani2018multimodal}, and KMLRR \cite{abavisani2018multimodal}. 

\begin{table}[t]
\begin{center}
\resizebox{\linewidth}{!}{%
\begin{tabular}{|l|c|c|c|}
\hline
\backslashbox{Structure}{Function} &	Max-pooling	& 	Additive	&	Concatenation\\
\hline
Early fusion &	$\times$	& 	$\times$	&	Early-concat. \\
\hline
Intermediate fusion &	Interm.-mpool.	& 	Interm.-additive	&	Interm.-concat. \\
\hline
Late fusion &	Late-mpool.	& 	Late-additive	&	Late-concat. \\
\hline
\end{tabular}
}
\caption{Spatial fusion variations that are used in the experiments. } \label{tbl:spatialmethods}
\end{center}
\end{table}

Also, to explore the contribution of leveraging information from multiple modalities into the performance of subspace clustering task, we report the performance of  subspace clustering methods on the single modalities as well.   In particular, we report the classical SSC~\cite{SSC_PAMI} and LRR~\cite{LRR_PAMI_2013} performances on the individual modalities along with the recently proposed DSC method \cite{deepsc17nips}.  Furthermore, we train an encoder-decoder similar to the network in \cite{deepsc17nips} but without the self-expressive layer, and extract the latent space representations.  These deep features are then fed to the SSC algorithm for clustering.   We call this method ``AE+SSC''.  This baseline will show the significance of using an end-to-end deep learning method for subspace clustering. 
  In our tables, we use boldface letters to denote the top performing method and specify the corresponding modalities or datasets in the rows, and subspace clustering methods on the columns. \\

\begin{table}[t!]
\begin{center}
\resizebox{\linewidth}{!}{%
\begin{tabular}{|c|l|c|c|c|c|}
\cline{3-6} 
\multicolumn{2}{c|}{ }& DSC\cite{deepsc17nips}  & AE+SSC & SSC\cite{SSC_PAMI}& LRR\cite{LRR_PAMI_2013} \\
\cline{3-6}\hline 
\multirow{3}{*}{MNIST}
        &   ACC   &  \textbf{92.05}  &  70.1  &  67.5  &  67.4  \\
      \cline{2-6}
      &NMI	 &	\textbf{87.07}	& 	80.94   &    71.64	& 	66.51    \\
            \cline{2-6}
      &ARI	 &	\textbf{84.60}	& 	62.33   &   57.03	& 	58.33    \\
\hline\hline
\multirow{3}{*}{USPS}
        &   ACC   &  \textbf{72.15}  &  69.9  &  37.5  &  44.35  \\
      \cline{2-6}
      &NMI	 &	74.73 	& 	\textbf{80.98}  &    36.61	& 	35.18    \\
            \cline{2-6}
      &ARI	 &	\textbf{65.47}	& 	62.41   &    28.40	& 	32.11    \\
\hline
\end{tabular}
}
\caption{The performance of single modality subspace clustering methods on Digits. Experiments are evaluated by average ACC, NMI and ARI over 5 runs. We use boldface for the top performer. Columns specify the single modality subspace clustering method, and rows specify the modality (MNIST or USPS) and criteria.} \label{tbl:digits_uni}
\end{center}
\end{table}

\noindent\textbf{Structures: }
We perform all the experiments on different datasets using the same protocol and network architectures to ensure fair and meaningful comparisons (including the networks for the single modality experiments).		 All the encoders have four convolutional layers, and decoders are stacked three deconvolution layers mimicking the inverse task of the encoder.  The network details are given in the Appendix.  

For the spatial fusion experiments, in the case of early fusion, we apply the fusion functions on the pixel intensities, and the rest of the network is similar to that of the single modality deep subspace clustering network.		   Conducted experiments for the intermediate fusion use a prior knowledge on the importance of the modalities.		They integrate weak modalities in the second hidden layer, and then, the combination of them in the third layer.		Finally, the fusion of all the weak modalities is combined with the strong modality (for example the visible domain in the ARL dataset) in the fourth layer.		   In the case of late fusion, all the modalities are fused in the fourth layer of the encoder.   

As discussed earlier, in the affinity fusion method there exists an encoder-decoder and a latent space per number of available modalities. For example, in the case of the ARL dataset with 5 modalities, we have 5 distinct encoders and decoders connected with a shared self-expressive layer. 		 For each modality in the experiments with the shared affinity, we use similar encoder-decoders as in the case of the DSC network \cite{deepsc17nips} with unimodal experiments.\\

\noindent\textbf{Training details: } We implemented our method in Python-2 with Tensorflow-1.4 \cite{abadi2016tensorflow}. We use the adaptive momentum-based gradient descent method (ADAM) \cite{kingma2014adam} to minimize our loss functions, and apply a learning rate of $10^{-3}$.

The input images of all the modalities are resized to $32\times32$, and rescaled to have pixel values between 0 and 255.  In our experiments, the Frobenius norm (i.e. $p=2$) is used in the loss functions \eqref{eq:DSC},~\eqref{eq:SFDMSC} and~\eqref{eq:DMSC} while training the networks.  Similar to~\cite{deepsc17nips}, for all the methods that have self-expressive layer, we start training on the specified objective functions in each model after a stage of pre-training on the dataset without the self-expressive layer.		  In particular, for all the proposed deep multimodal subspace clustering methods, and the unimodal DSC networks in the experiments with individual modalities, we pre-train the encoder-decoders for $20k$ epochs with the following objective
$$
\min_{\hat{\boldsymbol{\Theta}}} \sum_{m=1}^{M} \| \mathbf{X}^m -  \hat{\mathbf{X}}_{\hat{\boldsymbol{\Theta}}}^m\|^2_F,
$$ 
where $\hat{\boldsymbol{\Theta}}$ indicates the union of parameters in the encoder and decoder networks.		Note that for the unimodal experiments, $M=1$.

We use a batch size of $100$ for the pretraining stage of all the experiments.  However, once we start training the self-expressive layer, the method requires all the data points to be fed as a batch. Thus, in the experiments with digits, ARL faces and Yale-B facial components the batch sizes are 2000, 2160 and 2432, respectively.

We set the regularization parameters as $\lambda_1 =1$ and $\lambda_2 = 1 \times 10^{\frac{K}{10}-3}$,  where $K$ is the number of subjects in the dataset.   This experimental rule has been found to be efficient in \cite{deepsc17nips} as well.    A sensibility analysis over the range $[10^{-4},10^{4}]$ in Section~\ref{sec:hyperparameters}, shows that if $\lambda_1$ and $\lambda_2$ are kept around the same scale as our selections, the performance of the proposed method is not much sensitive to these parameters for a set of wide ranges. \\

\noindent\textbf{Evaluation metrics: } We compare the performance of different methods using the clustering accuracy rate (ACC),  normalized mutual information (NMI)~\cite{vinh2010information}, and Adjusted Rand Index (ARI)~\cite{rand1971objective} metrics.

In external validation of clustering methods where ground truth labels are available,  a correct clustering is usually referred as assigning objects belonging to the same category in the ground truth to the same cluster, and objects belonging to different categories to different clusters.  With that, ACC is defined as the number of data points correctly clustered divided by the total number of data points.  The ARI metric, in addition to penalizing the misclustered data points, penalizes putting two objects with the same label in different clusters, and is adjusted such that a random clustering will score close to 0.   The NMI captures the mutual information between the correct labels and the predicted labels, and is normalized between the range [0,1].

\subsection{Handwritten Digits}
In the first set of experiments, we use the 10 classes (i.e. digits) from the  MNIST  and the USPS datasets.  Figure~\ref{fig:datasets} (a) shows example images from these datasets.		       For the experiments with digits, we randomly sample 200 images per class from their training sets to reduce the computations and adjust the imbalance in the tests.		    

\begin{table*}[t!]
\begin{center}
\resizebox{\linewidth}{!}{%
\begin{tabular}{|c|l|c|c|c|c|c|c|c|}
\cline{3-9} 
\multicolumn{2}{c|}{ }&  CMVFC\cite{cao2015constrained} & TM-MSC\cite{zhang2015low}   &  MSSC\cite{abavisani2018multimodal} & MLRR\cite{abavisani2018multimodal} &KMSSC\cite{abavisani2018multimodal} & KMLRR\cite{abavisani2018multimodal} & Early-concat.\\
\cline{3-9}\hline
\multirow{3}{*}{Digits}
        &   ACC   &  47.6  &  80.65  &  81.65  &  80.6  &  84.4  &  86.85  &  92.2  \\
      \cline{2-9}
      &NMI	 &	73.56 	& 	83.44   &    85.33	& 	84.13  &	89.45 	& 	80.34   &    88.53	\\
            \cline{2-9}
      &ARI	 &	38.12 	& 	75.67   &    77.36	& 	76.53 	&	79.61 	& 	82.76   &    84.60	\\
\hline\hline
\multirow{3}{*}{ARL}
        &   ACC   &  96.58  &  96.64  &  97.78  &  97.5  &  97.97  &  97.74  &  98.24  \\
      \cline{2-9}
      &NMI	 &	98.39 	& 	98.35    &    99.58	& 	99.57  &	99.51 	& 	99.58   &    99.27	 \\
      \cline{2-9}
             &ARI	 &	94.85 	& 	95.85    &    96.40	& 	95.79  &	  96.09 	& 	95.88   &    97.21 \\
\hline\hline
\multirow{3}{*}{Extended Yale-B}
        &   ACC   &  66.84  &  63.12  &  80.3  &  67.62  &  87.65  &  82.45  &  65.55  \\      \cline{2-9}
      &NMI	 &	72.03	& 	67.06    &    82.78	& 	73.36   &	81.50	& 	85.43   &    78.82	\\
      \cline{2-9}
            &ARI	 &	40.00 	& 	38.37    &    50.18	& 	40.85  &	63.83 	& 	59.71   &    41.95	 \\
\hline \multicolumn{9}{c}{ }\\
\cline{3-9}
\multicolumn{2}{c|}{ }&   Interm.-concat.		& Interm-addition &  Interm.-mpool.& Late-concat.		& Late-addition & Late-mpool & Affinity fusion \\
\cline{3-9}\hline
\multirow{3}{*}{Digits}
      &  ACC	&	\texttt{\scriptsize{N/A}}	& 	\texttt{\scriptsize{N/A}}   &    \texttt{\scriptsize{N/A}}	&	91.15  &  \textbf{95.15}  &  91.45  &  \textbf{95.15}  \\
      \cline{2-9}
      &NMI	 &	\texttt{\scriptsize{N/A}}	& 	\texttt{\scriptsize{N/A}}    &    \texttt{\scriptsize{N/A}}	& 	84.28  &	91.35	& 	89.32   &    \textbf{92.09}	\\
      \cline{2-9}
            &ARI	 &	\texttt{\scriptsize{N/A}} 	& 	\texttt{\scriptsize{N/A}}    &   \texttt{\scriptsize{N/A}} 	& 	85.46  &	89.72 	& 	87.74   &    \textbf{90.22}	 \\
\hline\hline
\multirow{3}{*}{ARL}
        &   ACC   &  97.79  &  96.21  &  94.99  &  98.22  &  96.68  &  95.77  &  \textbf{98.34}  \\
      \cline{2-9}
      &NMI	 &	\textbf{99.59} 	& 98.95	   &    98.19	& 	99.31  &	99.23 	& 	98.92   &    99.36	   \\
      \cline{2-9}
            &ARI	 &	95.85 	& 	  94.64	& 	92.93  &	  97.02 	& 	96.24 	& 	94.77   &    \textbf{97.51} \\
\hline\hline
\multirow{3}{*}{Extended Yale-B}
        &   ACC   &  94.88  &  97.65  &  7.76  &  92.45  &  67.41  &  7.06  &  \textbf{99.22}  \\ 
      \cline{2-9}
      &NMI	 &	93.90 	& 	96.88    &   9.31	& 	92.53  &	66.95	& 	6.39   &    \textbf{98.89}	   \\
      \cline{2-9}
            &ARI	 &	88.19	& 	94.96   &    0.73	& 	82.91  &	33.37 	& 	00.48   &   \textbf{98.38}	 \\
\hline
\end{tabular}
}
\tiny{$\bullet$ \texttt{\scriptsize{N/A}} indicates that the corresponding method is not applicable to this experiment.}\hfill 
\caption{The performance of multimodal subspace clustering methods.		Each experiment is evaluated by average ACC,  NMI and ARI over 5 runs. We use boldface for the top performer. Columns of this table show the multimodal subspace clustering method, and the rows list datasets and clustering metrics. } \label{tbl:multi}
\end{center}
\end{table*}

We randomly bundle the same class samples across the two datasets and assume they present two modalities (views) of a digit.		One can see from Figure \ref{fig:datasets} (a), that the needed receptive field for recognizing the digits in the MNIST and the USPS datasets is relatively large.		  Based on this logic, in the experiments with digits, we use large kernels in the encoders.		The detailed network settings for these experiments are described in the Appendix.  Note that some structures including the late fusion methods in Table~\ref{tbl:spatialmethods} and the affinity fusion method  have more than one branches in some of their layers.

Table~\ref{tbl:digits_uni} shows the performance of deep subspace clustering per individual digits.		  This table reveals that the MNIST dataset is easier than the USPS dataset for the subspace clustering task.		 This observance coincides with the performance of other methods reported in ~\cite{guo2017deep}.		

Note that while the DSC method in Table~\ref{tbl:digits_uni} shows the-state-of-the-art performance on both datasets, a successful multimodal method should enhance the performance by leveraging the information across the two modalities.		   Table~\ref{tbl:multi} compares the performance of the multimodal methods in terms of accuracy, NMI and ARI metrics.		  We observe that most of the multimodal methods can successfully integrate the complementary information of the datasets in the subspace clustering task and provide a better performance in comparison to their unimodal counterpart.    However, the proposed deep multimodal subspace clustering methods perform significantly better than the classical multimodal subspace clustering methods.  In particular, the affinity fusion and late-addition methods can segment the digits with an accuracy of $95.15\%$, and NMI and ARI metric of above $90\%$.


\subsection{ARL Heterogeneous Face Dataset}
To test our methods on clustering datasets with a large number of subjects, we use the ARL dataset~\cite{hu2016polarimetric} which consists of facial images from 60 unique individuals in different spectrums and from different distances.		  This dataset has  facial images in the visible domain as well as  four different polarimetric thermal domains.		        Each subject has several well-aligned facial images per each modality.		         Sample images from this dataset are shown in Figure~\ref{fig:datasets} (b).  

Table~\ref{tbl:arl_uni} compares the performance of subspace clustering methods on individual modalities in the ARL dataset.		 As expected, the visible modality shows better performance among the different spectrums.	  As the samples are well-aligned in this dataset, we see that most of the subspace clustering methods work well across all the modalities.    In particular, the LRR method which takes the advantage of aligned data points, provides comparable results to the DSC method.

\begin{table}[t!]
\begin{center}
\resizebox{\linewidth}{!}{%
\begin{tabular}{|c|l|c|c|c|c|}
\cline{3-6} 
\multicolumn{2}{c|}{ }& DSC\cite{deepsc17nips}  & AE+SSC & SSC\cite{SSC_PAMI}& LRR\cite{LRR_PAMI_2013} \\
\cline{3-6}\hline
\multirow{3}{*}{Visible}
        &   ACC   &  \textbf{92.54}  &  89.87  &  81.86  &  91.07  \\
      \cline{2-6}
      &NMI	 &	\textbf{97.03} 	& 	96.25    &    94.56	& 	97.16    \\
            \cline{2-6}
      &ARI	 &	\textbf{92.54} 	& 	88.08    &    72.32	& 	89.94    \\
\hline\hline
\multirow{3}{*}{DP}
        &   ACC   &  \textbf{91.81}  &  89.08  &  63.2  &  89.4  \\
              \cline{2-6}
      &NMI	 &	  \textbf{97.60}	& 	97.17   & 83.59	&	95.71	\\
                  \cline{2-6}
      &ARI	 &	\textbf{91.69} 	& 	87.48    &    47.98	& 	85.47    \\
      \hline\hline
\multirow{3}{*}{S0}
        &   ACC   &  \textbf{62.64}  &  55.38  &  21.58  &  57.23  \\
      \cline{2-6}
      &NMI	 &	\textbf{84.20} 	& 	77.62    &    47.83	& 	80.44    \\
                  \cline{2-6}
      &ARI	 &	\textbf{49.23} 	& 	41.60    &    11.63	& 	36.56    \\
      \hline\hline
\multirow{3}{*}{S1}
        &   ACC   &  \textbf{91.72}  &  86.21  &  54.68  &  86.12  \\
      \cline{2-6}
      &NMI	 &	\textbf{97.09} 	& 	96.55    &    78.60	& 	95.13    \\
                  \cline{2-6}
      &ARI	 &	\textbf{89.55} 	& 	86.16   &    42.69	& 	85.62    \\
      \hline\hline
\multirow{3}{*}{S2}
        &   ACC   &  \textbf{89.68}  &  89.26  &  57.92  &  85.88  \\
      \cline{2-6}
      &NMI	 &	  \textbf{97.63}	& 	97.38   & 82.77	&	94.73	\\

            \cline{2-6}
      &ARI	 &	\textbf{89.34} 	& 	88.05    &    43.38	& 	84.05    \\
            \hline
\end{tabular}
}
\caption{The performance of single modality subspace clustering methods on ARL dataset. Experiments are evaluated by average ACC, NMI and ARI over 5 runs. We use boldface for the top performer. Columns specify the single modality subspace clustering method, and rows specify the modalities and criteria.} \label{tbl:arl_uni}
\end{center}
\end{table}

Since the ARL dataset has multiple modalities, beside the early and late fusion structures, we also use an intermediate structure when designing the multimodal encoders.		  Hence, in this experiment, we add the following intermediate spatial fusion structure to the multimodal methods.		 Assuming the visible domain is the main modality, we integrate $S0$, $S1$ and $S2$ modalities in the second layer and combine their fused output with the $DP$ samples in the third layer.		  Finally, we fuse the result with the visible domain at the last layer of the encoders.

The performances of deep multimodal subspace clustering methods are compared in Table~\ref{tbl:multi}.		 We observe that most of the methods are able to leverage the complementary information of the different spectrums and provide a more accurate clustering in comparison to the unimodal performances.		  In particular, the affinity fusion method has the best performance, and late-concat and early-concat methods provide comparable results.		This experiment clearly shows that our proposed methods can perform well even with a large number of subjects in the dataset.

\begin{figure}[t]
 \centering \includegraphics[width=.21\textwidth]{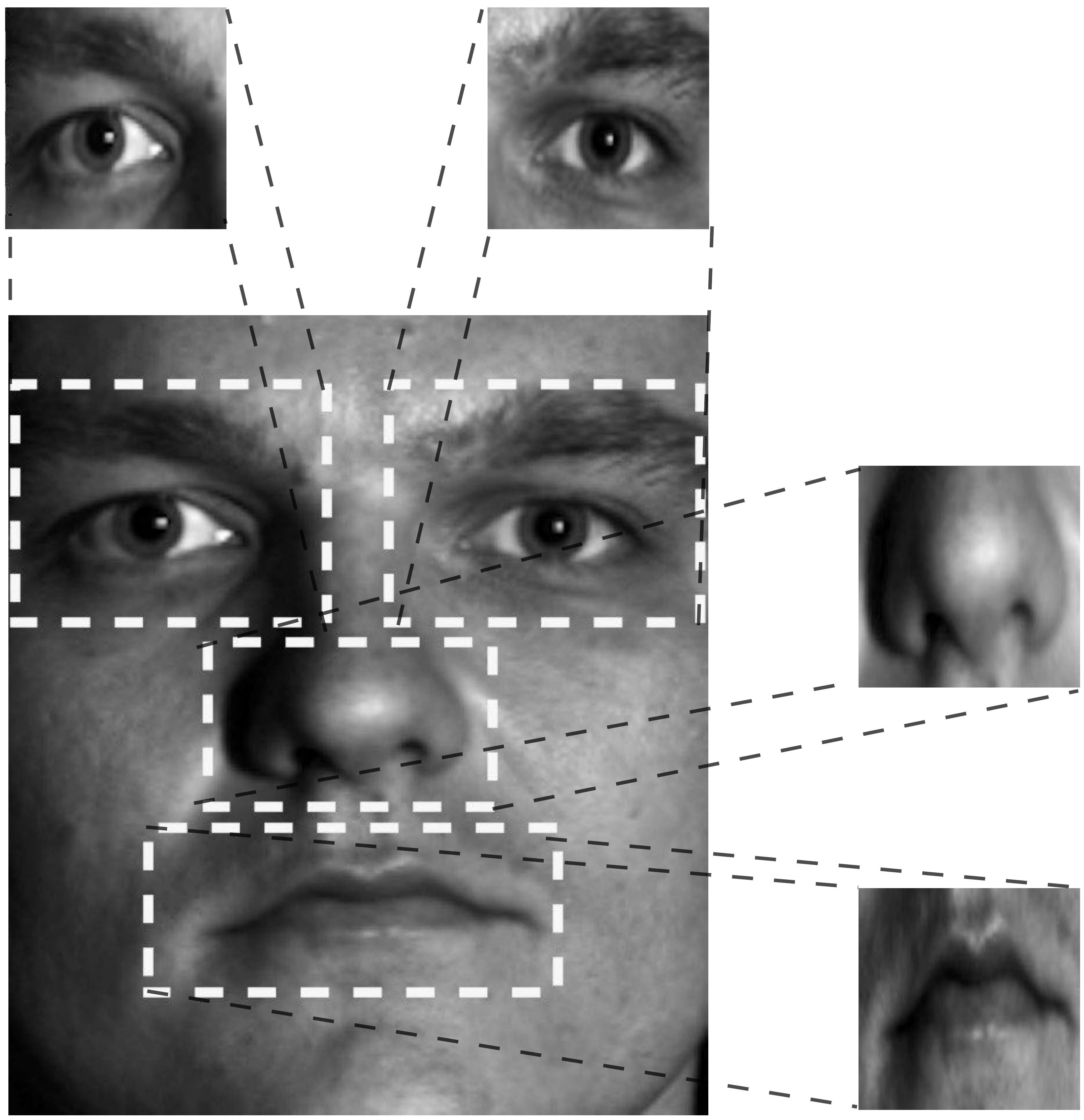}
\vskip -10pt\caption{Facial components are extracted by applying a fixed mask on the faces in the Extended Yale-B dataset~\cite{9ptsLight}.}
\label{fig:mask}
\end{figure}


\begin{table}[t!]
\begin{center}
\resizebox{\linewidth}{!}{%
\begin{tabular}{|c|l|c|c|c|c|}
\cline{3-6} 
\multicolumn{2}{c|}{ }& DSC\cite{deepsc17nips}  & AE+SSC & SSC\cite{SSC_PAMI}& LRR\cite{LRR_PAMI_2013} \\
\cline{3-6}\hline
\multirow{3}{*}{Face}
        &   ACC   &  \textbf{96.82}  &  72.93  &  72.78  &  63.34  \\ 
      \cline{2-6}
      &NMI	 &	\textbf{94.82} 	& 	79.10   &    79.17	& 	70.08    \\
       \cline{2-6}
            &ARI	 &	\textbf{91.31} 	& 	43.94   &    42.90	& 	37.38    \\
\hline\hline
\multirow{3}{*}{Right-eye}
        &   ACC   &  \textbf{87.62}  &  83.34  &  66.84  &  65.35  \\ 
      \cline{2-6}
      &NMI	 &	\textbf{89.19} 	& 	86.99    &    73.62	& 	69.33    \\
       \cline{2-6}
      &ARI	 &	\textbf{75.05} 	& 	61.90   &    39.66	& 	38.37    \\
\hline\hline
\multirow{3}{*}{Left-eye}
        &   ACC   &  \textbf{80.94}  &  72.24  &  63.02  &  63.08  \\ 
      \cline{2-6}
      &NMI	 &	\textbf{79.58}	& 	76.48    &    69.08	& 	70.13   \\
       \cline{2-6}
      &ARI	 &	\textbf{50.17} 	& 	42.94   &    33.12	& 	34.07    \\
\hline\hline
\multirow{3}{*}{Nose}
        &   ACC   &  \textbf{67.53}  &  51.61  &  41.51  &  39.9  \\
              \cline{2-6}
      &NMI	 &	\textbf{75.23}	& 	61.64    &    50.78	& 	48.73   \\
       \cline{2-6}
      &ARI	 &	\textbf{40.82} 	& 	22.96   &    16.67	& 	15.13  \\
\hline\hline
\multirow{3}{*}{Mouth}
        &   ACC   &  \textbf{76.86}  &  67.42  &  56.07  &  62.92  \\
              \cline{2-6}
      &NMI	 &	\textbf{76.42}	& 	72.91    &    64.11	& 	67.28    \\
       \cline{2-6}
      &ARI	 &	\textbf{43.90} 	& 	40.52   &    25.71	& 	33.02    \\
\hline
\end{tabular}
}
\caption{The performance of single modality subspace clustering methods on Extended Yale-B dataset. Experiments are evaluated by average ACC, NMI and ARI over 5 runs. We use boldface for the top performer. Columns specify the single modality subspace clustering method, and rows specify the facia components and criteria. } \label{tbl:yaleb_uni}
\end{center}
\end{table}

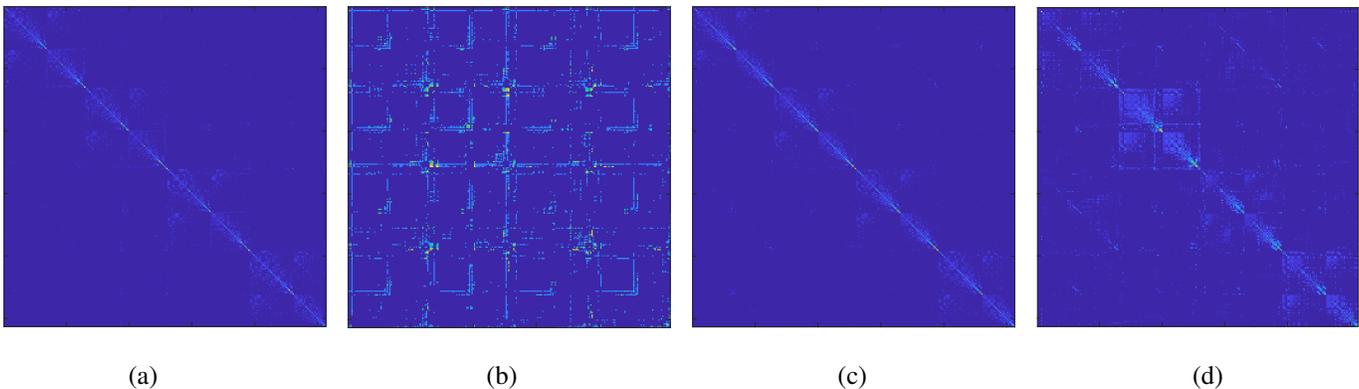
\begin{figure*}[t]
\centering   \begin{overpic}[width=1\textwidth,tics=5]{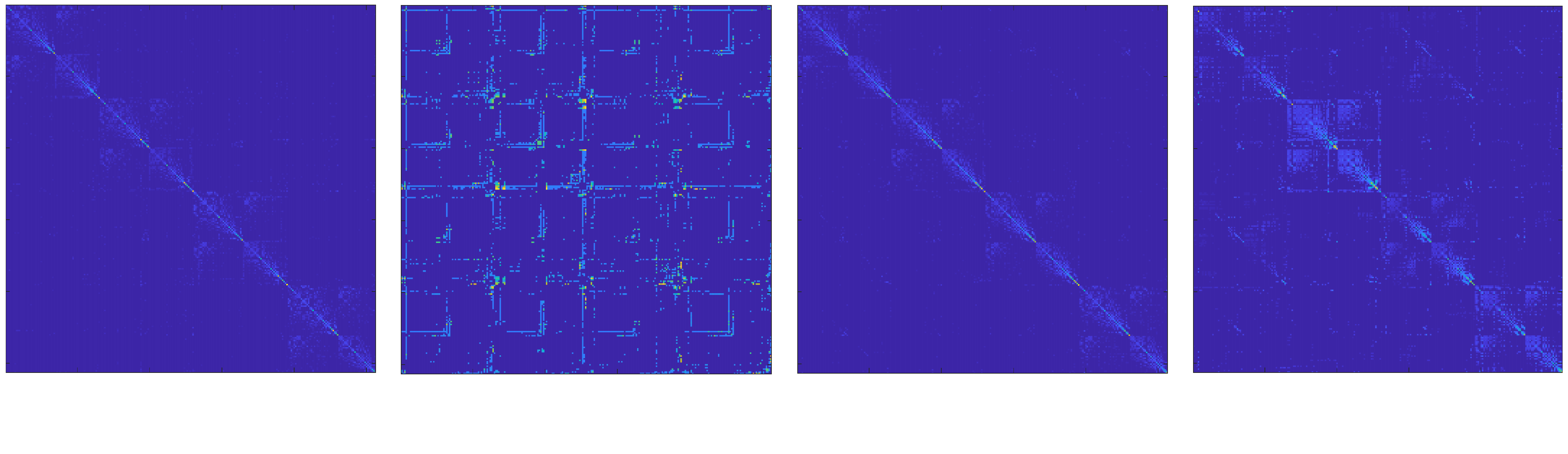}
\put (9.5,1) {{(a)}}
\put (35.75,1) {{(b)}}
\put (61.5,1) {{(c)}}
\put (87.5,1) {{(d)}}
\end{overpic}
\vskip -15pt\caption{Visualization of the affinity matrices for first four subjects in the Extended Yale-B dataset calculated from the self-expressive layer weight matrices in		  (a) unimodal clustering on faces using DSC.		(b) The \textit{late-mpool} method.		(c) The \textit{late-concat} method.	(d) The \textit{affinity fusion} method.  Note that (b) shows a failure case of the spatial fusion methods.}
\label{fig:compareaffinities}
\end{figure*}

\subsection{Facial Components}
The Extended Yale-B dataset~\cite{9ptsLight} consists of 64 frontal images of $38$ individuals under varying illumination conditions.		  This dataset is popular in subspace clustering studies~\cite{deepsc17nips,LRR_PAMI_2013,SSC_PAMI}.		   We crop the facial components (i.e. eyes, nose and mouth), and view them as weak modalities.  In the biometrics literature, they are viewed as soft biometrics \cite{MSRC}.		  To crop the facial components, we apply a fixed face mask as shown in Figure~\ref{fig:mask} on all the facial images.	The extracted facial regions are resized to $32\times32$ images.    This experiment is especially important as the modalities do not share the spatial correspondence.		For example, spatial locations in the mouth modality cannot be projected on the spatial positions in the nose modality.	 Sample images from this dataset are shown in Figure~\ref{fig:datasets} (c).	 The setting in this experiment can examine the proposed methods under the condition of spatially unrelated modalities.		

The performance of subspace clustering methods on the individual facial components is summarized in Table~\ref{tbl:yaleb_uni}.		 We observe that the nose and the mouth modalities fail to provide good clustering results. On the other hand, DSC and AE+SSC perform well on the eye and the entire face modalities.

Since the mouth, nose, and eyes are considered as weak modalities, in the design of the intermediate spatial fusion we combine the two eyes, and the mouth and the nose separately in the second layer of the encoders, and fuse the result of their combinations in the third layer.		Finally, we fuse the combined features with the face features in the fourth layer.

The performance of various multimodal subspace clustering methods are tabulated in Table~\ref{tbl:multi}.	  It is worth highlighting several interesting observations from the results.   As can be seen, the \textit{late-mpool} and \textit{interm-mpool} methods fail to segment the data points.  	That is because this fusion function at each spatial position returns the maximum of the activation values at the same spatial position between its input feature maps.  Since the modalities do not share any spatial correspondence in this experiment, this function does not provide good performance.   In addition,  even though additive and concatenate fusion functions have provided good results in some cases, because of a similar reason their performances are highly related to the structure choices.  For example, the additive function provides better performance with the intermediate fusion structure, while the concatenation works better with the late fusion structure choice.     However,  the \textit{affinity fusion} provides the state-of-the-art clustering performance of above $99\%$ accuracy, the NMI of $98.89\%$ and ARI metric of $98.38\%$.   This is mainly due to the fact that this method does not rely on the spatial correspondence among the modalities. 
 
Figure~\ref{fig:compareaffinities} compares the affinity matrices of the first four subjects in the Extended Yale-B datasets.  The affinity matrices are calculated from the self-expressive layer weights of their corresponding trained networks.      The depicted affinity matrices in these figures are the result of a permutation being applied on the matrix so that data points of the same clusters are alongside each other.   With this arrangement, a perfect affinity matrix should be block diagonal. 

Figure~\ref{fig:compareaffinities} (a) shows the affinity matrix corresponding to the DSC method for clustering faces.   Figure~\ref{fig:compareaffinities} (b) shows this matrix for the multimodal subspace clustering with the \textit{late-mpool} method.  Note that this method fails to cluster the data, and as can be seen, its affinity matrix is not block-diagonal.      Figure~\ref{fig:compareaffinities} (c) and Figure~\ref{fig:compareaffinities} (d) show the affinity matrices of the \textit{late-concat} and \textit{affinity fusion} methods, respectively.  We observe that both methods provide a solid block diagonal affinity matrices.

\subsection{Convergence study}
\begin{figure}[t]
\centering   \begin{overpic}[width=0.35\textwidth,tics=5]{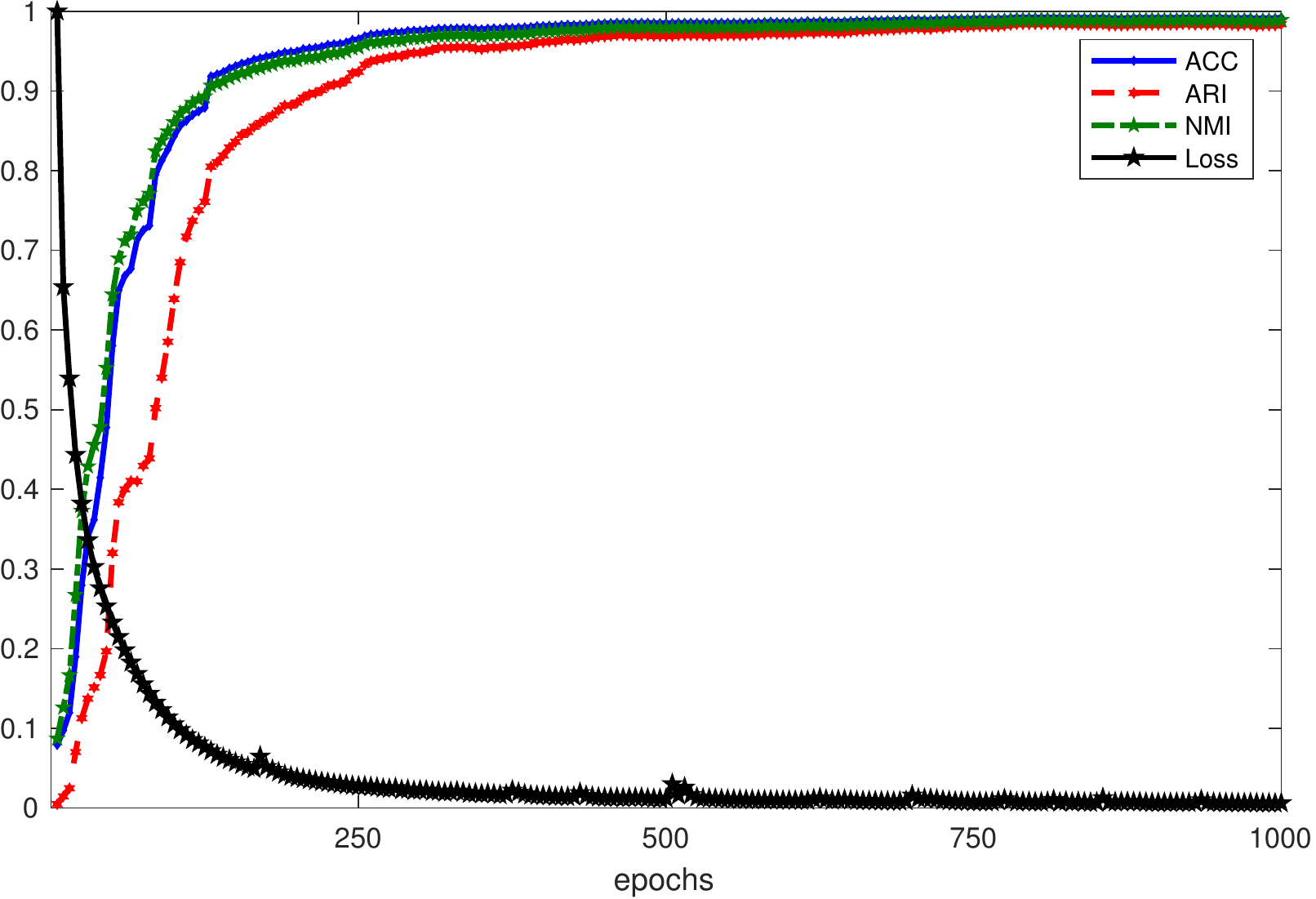}
\end{overpic}
\vskip -5pt\caption{The \emph{affinity fusion} method's loss function and the clustering metrics over different training epochs in the Yale-B facial components experiment.   The reported values in this figure are normalized between zero and one. This figure shows the convergence of our objective function.}
\label{fig:convergence}
\end{figure}

To empirically show the convergence of our proposed method, in Figure~\ref{fig:convergence}, we show the objective function of the \emph{affinity fusion} method and its clustering metrics vs iteration plot for solving \eqref{eq:DMSC}.  The reported values in Figure~\ref{fig:convergence} are normalized between zero and one.   As can be seen from the figure, our algorithm converges in a few iterations.

\subsection{Regularization parameters}\label{sec:hyperparameters}
\begin{figure}[t]
\centering   \begin{overpic}[width=0.49\textwidth,tics=5]{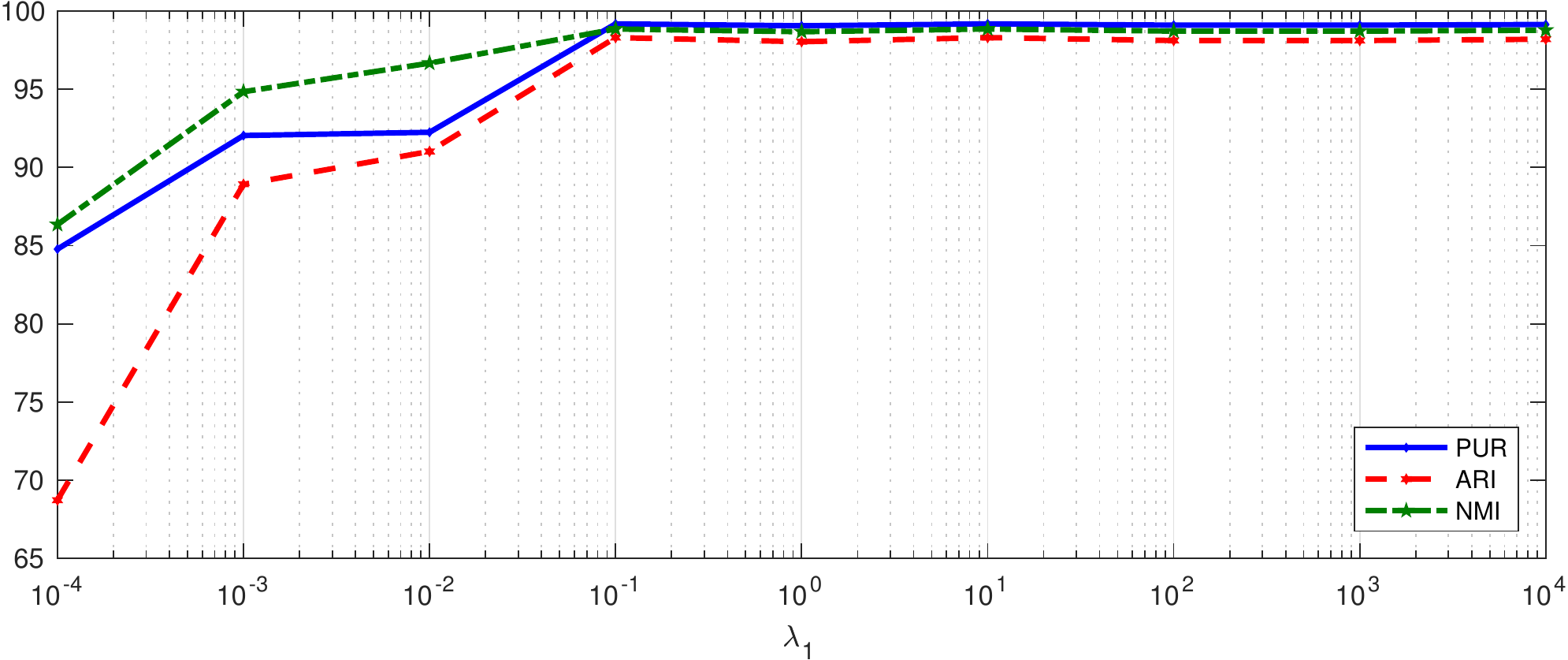}
\end{overpic}\\
\bigskip
\centering   \begin{overpic}[width=0.49\textwidth,tics=5]{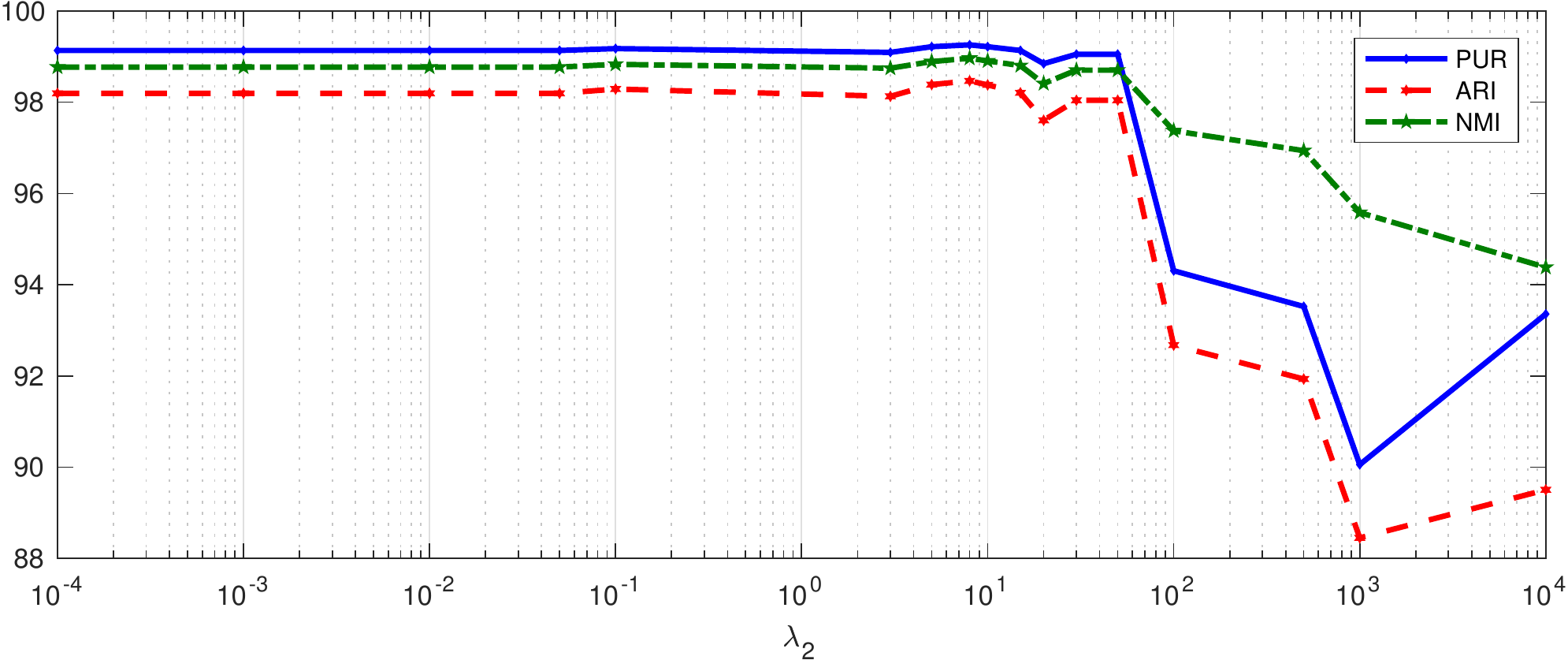}
\end{overpic}
\vskip -10pt\caption{The \emph{affinity fusion} method's performance through different parameter selections for $\lambda_1$ and $\lambda_2$.}
\label{fig:lambdas}
\end{figure}

In this section, we analyze the sensibility of the proposed method to the regularization parameters $\lambda_1$ and $\lambda_2$ in the loss function~\eqref{eq:DMSC}.   Figure~\ref{fig:lambdas} shows the influence of these regularization parameters on the performance of the \emph{affinity fusion} method on the Extended Yale-B dataset.   

In Figure~\ref{fig:lambdas} (a), we fix $\lambda_2 = 1$ and report the metrics with various  $\lambda_1$s over the range of $[10^{-4},10^{4}]$.     Similarly, in Figure~\ref{fig:lambdas} (b), we fix $\lambda_1 = 1$  and this time change $\lambda_2$ in the similar range to analyze the influence of $\lambda_2$ on the performance of the method.  As can be seen from the figure, in a wide range of values, the final performance of the method is not sensitive to the choice of parameters.  The experimental setting suggested in \cite{deepsc17nips} also performed well in all the experiments. 

\subsection{Performance with respect to different norms on the self-expressive layer}
In this section, we compare the performance of the proposed \emph{affinity fusion} method by changing the  $p$-norm on the self-expressive layer in the optimization problem~\eqref{eq:DMSC}.   Table~\ref{tbl:norms} reports the clustering metrics for the experiments with $p=0.3$, $p=1$, $p=1.5$ and $p=2$.   As can be seen from this table, while experiments with $p=1$, $p=1.5$ and $p=2$ have comparable performances, applying the p-norm with $p=0.3$ does not provide sufficient result.  It is worth mentioning that in our experiments with different norms with $0.3<p<1$ the method showed instability, and for $p<0.3$ the minimization of~\eqref{eq:DMSC} did not converge.  The reason is that the norms with $p<1$ are non-convex, and one might need additional regularizations to keep the optimization tractable. 

\begin{table}[t]
\begin{center}
\resizebox{\linewidth}{!}{%
\begin{tabular}{|l|c|c|c|c|c|c|}
\hline
\backslashbox{Metric}{$\|\cdot\|_{p}$} &	$p<0.3$	& 	$p=0.3$	& $p = 1$ & $p = 1.5$ & $p = 2$ \\
\hline
PUR &	$\times$	& 	09.32 &	99.13 &	99.17	& 	\textbf{99.22} \\
\hline
NMI &	$\times$	& 	18.64	&	98.78&	98.84	& 	\textbf{98.89}	\\
\hline
ARI &	$\times$	& 	02.38	&	98.20&	98.29	& 	\textbf{98.38}\\
\hline
\end{tabular}
}
\caption{Analysis of different regularization norms on the self-expressive layer.  Our experiments with $p<0$ did not converged. The results are 5-fold average. We use boldface for the top performer.} \label{tbl:norms}
\end{center}
\end{table}

\section{Conclusion}\label{sec:con}
We presented novel deep multimodal subspace clustering networks for clustering multimodal data.  In particular, we presented two fusion techniques of spatial fusion and affinity fusion.   We observed that spatial fusion methods in a deep multimodal subspace clustering task relay on spatial correspondences among the modalities.  On the other hand, the proposed affinity fusion that finds a shared affinity across all the modalities provides the state-of-the-art results in all the conducted experiments.  This method clusters the images in the Extended Yale-B dataset with an accuracy  of $99.22\%$, normalized mutual information of $98.89\%$ and adjusted rand index of $98.38\%$.

\section*{Appendix: Network Architectures}\label{sec:appendix}
In this section, we provide the details of the network architecture used in the experiments.  Note that all the plugged in convolutional layers use  \textit{relu} as well. 

\subsection{Different networks corresponding to digits experiments}

\begin{table}[htp!]
	\centering
	\caption{Early-fusion networks in the digits experiments.}
	\label{table:tbl3r}
	\resizebox{\linewidth}{!}{
		\begin{tabular}{|c | c | c|c | l | p{0.7cm}|}
			\cline{2-6}
			\multicolumn{1}{c|}{ } & \multirow{3}{*}{Layer} & \multirow{3}{*}{Input} & \multirow{3}{*}{output} & \multirow{3}{*}{Kernel} & (stride, pad)  \\
			\cline{2-6}\hline			
			\multirow{3}{*}{Feature Fusion}  & \multirow{3}{*}{Fusion $1$} &  Image 1  & \multirow{3}{*}{Fusion $1$}  & \centering \multirow{3}{*}{-} & \multirow{3}{*}{-} \\
			&   & Image 2 &    &    &   \\	
			\hline\hline
			\multirow{4}{*}{Convolutional layers}  & Conv 1 &  Fusion $1$  & Conv 1  &  $1\times7\times7\times7$ & (2,1) \\
			&    Conv 2 &  Conv 1  & Conv 2  &  $1\times5\times5\times10$ & (2,1) \\			
			&    Conv 3 &  Conv 2  & Conv 3  &  $1\times3\times3\times15$ & (1,0) \\	
			&    Conv 4 &  Conv 3  & Latent  &  $1\times1\times1\times15$ & (1,0) \\
			\hline\hline
			\multirow{1}{*}{Self-expressiveness}  & \multirow{1}{*}{ $\Theta_s$} &  Latent    & L-recon  & \centering \multirow{1}{*}{$4000000$ Parameters} & \multirow{1}{*}{-} \\
			\hline\hline
			\multirow{1}{*}{Multimodal}  & \multirow{1}{*}{Decoder} &  \multirow{3}{*}{L-recon}    & Recon 1  & \centering \multirow{1}{*}{Details in} &  \\
			\multirow{1}{*}{Decoder} &\multirow{1}{*}{layers}&  & Recon 2  & \multirow{1}{*}{Table~\ref{tbl:decoder_digits}}& \\
			\hline						
		\end{tabular} 
	}
\end{table}

\begin{table}[htp!]
	\centering
	\caption{Late-fusion networks in the digits experiments.}
	\label{table:tbl3r}
	\resizebox{\linewidth}{!}{
		\begin{tabular}{|c|c|c|c|l|  p{0.7cm}|}
			\cline{2-6}
			\multicolumn{1}{c|}{ } & \multirow{3}{*}{Layer} & \multirow{3}{*}{Input} & \multirow{3}{*}{output} & \multirow{3}{*}{Kernel} & (stride, pad)  \\
			\cline{2-6}\hline	
			\multirow{4}{*}{Branch 1}  & B1/Conv  1 &  Image $1$  & B1/Conv 1  &  $1\times7\times7\times7$ & (2,1) \\
			&    B1/Conv 2 &  B1/Conv 1  & B1/Conv 2  &  $1\times5\times5\times10$ & (2,1) \\			
			&    B1/Conv 3 &  B1/Conv 2  & B1/Conv 3  &  $1\times3\times3\times15$ & (1,0) \\	
			&    B1/Conv 4 &  B1/Conv 3  & B1/out  &  $1\times1\times1\times15$ & (1,0) \\
			\hline\hline
			\multirow{4}{*}{Branch 2}  & B2/Conv 1 &  Image $2$  & B2/Conv 1  &  $1\times7\times7\times7$ & (2,1) \\
			&    B2/Conv 2 &  B2/Conv 1  & B2/Conv 2  &  $1\times5\times5\times10$  & (2,1) \\			
			&    B2/Conv 3 &  B2/Conv 2  & B2/Conv 3  &  $1\times3\times3\times15$ & (1,0) \\	
			&    B2/Conv 4 &  B2/Conv 3  & B2/out  &  $1\times1\times1\times15$ & (1,0) \\
			\hline\hline		
			
			\multirow{2}{*}{Feature Fusion}  & \multirow{2}{*}{Fusion $1$} &  B1/out   & \multirow{2}{*}{Latent}  & \centering \multirow{2}{*}{-} & \multirow{2}{*}{-} \\
			&   &  B2/out &    &    &   \\
			
			\hline\hline
			\multirow{1}{*}{Self-expressiveness}  & \multirow{1}{*}{ $\Theta_s$} &  Latent    & L-recon  & \centering \multirow{1}{*}{$4000000$ Parameters} & \multirow{1}{*}{-} \\
			\hline\hline
			\multirow{1}{*}{Multimodal}  & \multirow{1}{*}{Decoder} &  \multirow{3}{*}{L-recon}    & Recon 1  & \centering \multirow{1}{*}{Details in} & \multirow{3}{*}{} \\
			\multirow{1}{*}{Decoder} &\multirow{1}{*}{layers}&  & Recon 2  & \multirow{1}{*}{Table~\ref{tbl:decoder_digits}}& \\
			\hline	
		\end{tabular} 
	}
\end{table}

\begin{table}[htp!]
	\centering
	\caption{Affinity fusion networks in the digits experiments.}
	\label{table:tbl3r}
	\resizebox{\linewidth}{!}{
		\begin{tabular}{|c|c|c|c|l|  p{0.7cm}|}
			\cline{2-6}
			\multicolumn{1}{c|}{ } & \multirow{3}{*}{Layer} & \multirow{3}{*}{Input} & \multirow{3}{*}{output} & \multirow{3}{*}{Kernel} & (stride, pad)  \\
			\cline{2-6}\hline	
			\multirow{4}{*}{Encoder 1}  & B1/Conv  1 &  Image $1$  & B1/Conv 1  &  $1\times7\times7\times7$ & (2,1) \\
			&    B1/Conv 2 &  B1/Conv 1  & B1/Conv 2  &  $1\times5\times5\times10$ & (2,1) \\			
			&    B1/Conv 3 &  B1/Conv 2  & B1/Conv 3  &  $1\times3\times3\times30$ & (1,0) \\	
			&    B1/Conv 4 &  B1/Conv 3  & Latent 1  &  $1\times3\times3\times30$ & (1,0) \\
			\hline\hline
			\multirow{4}{*}{Encoder 2}  & B2/Conv 1 &  Image $2$  & B2/Conv 1  &  $1\times7\times7\times7$ & (2,1) \\
			&    B2/Conv 2 &  B2/Conv 1  & B2/Conv 2  &  $1\times5\times5\times10$  & (2,1) \\			
			&    B2/Conv 3 &  B2/Conv 2  & B2/Conv 3  &  $1\times3\times3\times15$ & (1,0) \\	
			&    B2/Conv 4 &  B2/Conv 3  & Latent 2  &  $1\times3\times3\times15$ & (1,0) \\
			\hline\hline		
			
			\multirow{1}{*}{Self-expressiveness}  & \multirow{2}{*}{Common $\Theta_s$} &  Latent 1   & L-recon 1  & \centering \multirow{2}{*}{$4000000$ Parameters} & \multirow{2}{*}{-} \\
			\multirow{1}{*}{layer}&   &  Latent 2 &  L-recon 2  &    &   \\
			
			\hline\hline	
			\multirow{3}{*}{Decoder 1}  & D1/deconv 1 &  L-recon $1$  & D1/deconv 1  & $1\times3\times3\times15$  & (1,0) \\
			&    D1/deconv 2 &  D1/deconv 1  & D1/deconv 2  &  $1\times5\times5\times10$ & (2,1) \\			
			&    D1/deconv 3 &  D1/deconv 2  & Recon 1  &  $1\times7\times7\times7$ & (2,1) \\	
			\hline\hline
			\multirow{3}{*}{Decoder 2}  & D2/deconv 2 &  L-recon $2$  & D2/deconv 1  & $1\times3\times3\times15$  & (1,0) \\
			&    D2/deconv 2 &  D2/deconv 1  & D2/deconv 2  &  $1\times5\times5\times10$ & (2,1) \\			
			&    D2/deconv 3 &  D2/deconv 2  & Recon 2  &  $1\times7\times7\times7$ & (2,1) \\	
			\hline
			
		\end{tabular} 
	}
\end{table}

\begin{table}[htp!]
	\centering
	\caption{Multimodal decoder in the digits experiments.}
	\label{tbl:decoder_digits}
	\resizebox{\linewidth}{!}{
		\begin{tabular}{|c|c|c|c|l|  p{0.7cm}|}
			\cline{2-6}
			\multicolumn{1}{c|}{ } & \multirow{3}{*}{Layer} & \multirow{3}{*}{Input} & \multirow{3}{*}{output} & \multirow{3}{*}{Kernel} & (stride, pad)  \\
			\cline{2-6}\hline	
			\multirow{3}{*}{Decoder 1}  & D1/deconv 1 &  L-recon  & D1/deconv 1  & $1\times3\times3\times15$  & (1,0) \\
			&    D1/deconv 2 &  D1/deconv 1  & D1/deconv 2  &  $1\times5\times5\times10$ & (2,1) \\			
			&    D1/deconv 3 &  D1/deconv 2  & Recon 1  &  $1\times7\times7\times7$ & (2,1) \\	
			\hline\hline
			\multirow{3}{*}{Decoder 2}  & D2/deconv 2 &  L-recon   & D2/deconv 1  & $1\times3\times3\times15$  & (1,0) \\
			&    D2/deconv 2 &  D2/deconv 1  & D2/deconv 2  &  $1\times5\times5\times10$ & (2,1) \\			
			&    D2/deconv 3 &  D2/deconv 2  & Recon 2  &  $1\times7\times7\times7$ & (2,1) \\	
			\hline
			
		\end{tabular} 
	}
\end{table}


\pagebreak

\subsection{Different networks corresponding to ARL experiments}

\begin{table}[htp!]
	\centering
	\caption{Early-fusion networks in the ARL experiments.}
	\label{table:tbl3r}
	\resizebox{\linewidth}{!}{
		\begin{tabular}{|c | c | c|c | l | p{0.7cm}|}
			\cline{2-6}
			\multicolumn{1}{c|}{ } & \multirow{3}{*}{Layer} & \multirow{3}{*}{Input} & \multirow{3}{*}{output} & \multirow{3}{*}{Kernel} & (stride, pad)  \\
			\cline{2-6}\hline			
			\multirow{5}{*}{Feature Fusion}  & \multirow{5}{*}{Fusion $1$} &  Image 1  & \multirow{5}{*}{Fusion $1$}  & \centering \multirow{5}{*}{-} & \multirow{5}{*}{-} \\
			&   & Image 2 &    &    &   \\
			&   & Image 3 &    &    &   \\
			&   & Image 4 &    &    &   \\
			&   & Image 5 &    &    &   \\	
			\hline\hline
			\multirow{4}{*}{Convolutional layers}  & Conv 1 &  Fusion $1$  & Conv 1  &  $1\times3\times3\times5$ & (2,1) \\
			&    Conv 2 &  Conv 1  & Conv 2  &  $1\times1\times1\times7$ & (2,1) \\			
			&    Conv 3 &  Conv 2  & Conv 3  &  $1\times1\times1\times15$ & (1,0) \\	
			&    Conv 4 &  Conv 3  & Latent  &  $1\times1\times1\times15$ & (1,0) \\
			\hline\hline
			
			\multirow{1}{*}{Self-expressiveness}  & \multirow{1}{*}{ $\Theta_s$} &  Latent    & L-recon  & \centering \multirow{1}{*}{$4665600$ Parameters} & \multirow{1}{*}{-} \\
			\hline\hline
			\multirow{5}{*}{Multimodal}  & \multirow{5}{*}{Decoder} &  \multirow{5}{*}{L-recon}    & Recon 1  & \centering \multirow{5}{*}{Details in} & \multirow{5}{*}{} \\
			\multirow{5}{*}{Decoder} &\multirow{5}{*}{layers}&  & Recon 2  & \multirow{5}{*}{Table~\ref{tbl:decoder_arl}}& \\
			&  &  & Recon 3  &  & \\
			&  &  & Recon 4  &  & \\
			&  &  & Recon 5  &  & \\	
			\hline	
			
		\end{tabular} 
	}
\end{table}

\begin{table}[htp!]
	\centering
	\caption{Late-fusion networks in the ARL experiments.}
	\label{table:tbl3r}
	\resizebox{\linewidth}{!}{
		\begin{tabular}{|c|c|c|c|l|  p{0.7cm}|}
			\cline{2-6}
			\multicolumn{1}{c|}{ } & \multirow{3}{*}{Layer} & \multirow{3}{*}{Input} & \multirow{3}{*}{output} & \multirow{3}{*}{Kernel} & (stride, pad)  \\
			\cline{2-6}\hline	
			\multirow{4}{*}{Branch 1}  & B1/Conv  1 &  Image $1$  & B1/Conv 1  &  $1\times3\times3\times5$ & (2,1) \\
			&    B1/Conv 2 &  B1/Conv 1  & B1/Conv 2  &  $1\times1\times1\times7$ & (2,1) \\			
			&    B1/Conv 3 &  B1/Conv 2  & B1/Conv 3  &  $1\times1\times1\times15$ & (1,0) \\	
			&    B1/Conv 4 &  B1/Conv 3  & B1/out  &  $1\times1\times1\times15$ & (1,0) \\
			\hline\hline
			\multirow{4}{*}{Branch 2}  & B2/Conv 1 &  Image $2$  & B2/Conv 1  &  $1\times3\times3\times5$ & (2,1) \\
			&    B2/Conv 2 &  B2/Conv 1  & B2/Conv 2  &  $1\times1\times1\times7$  & (2,1) \\			
			&    B2/Conv 3 &  B2/Conv 2  & B2/Conv 3  &  $1\times1\times1\times15$ & (1,0) \\	
			&    B2/Conv 4 &  B2/Conv 3  & B2/out  &  $1\times1\times1\times15$ & (1,0) \\
			\hline\hline		
			
			\multirow{4}{*}{Branch 3}  & B3/Conv 1 &  Image $3$  & B3/Conv 1  &  $1\times3\times3\times5$ & (2,1) \\
			&    B3/Conv 2 &  B3/Conv 1  & B3/Conv 2  &  $1\times1\times1\times7$  & (2,1) \\			
			&    B3/Conv 3 &  B3/Conv 2  & B3/Conv 3  &  $1\times1\times1\times15$ & (1,0) \\	
			&    B3/Conv 4 &  B3/Conv 3  & B3/out   &  $1\times1\times1\times15$ & (1,0) \\
			\hline\hline	
			\multirow{4}{*}{Branch 4}  & B4/Conv 1 &  Image $4$  & B4/Conv 1  &  $1\times3\times3\times5$ & (2,1) \\
			&    B4/Conv 2 &  B4/Conv 1  & B4/Conv 2  &  $1\times1\times1\times7$  & (2,1) \\			
			&    B4/Conv 3 &  B4/Conv 2  & B4/Conv 3  &  $1\times1\times1\times15$ & (1,0) \\	
			&    B4/Conv 4 &  B4/Conv 3  & B4/out   &  $1\times1\times1\times15$ & (1,0) \\
			\hline\hline		
			\multirow{4}{*}{Branch 5}  & B5/Conv 1 &  Image $5$  & B5/Conv 1  &  $1\times3\times3\times5$ & (2,1) \\
			&    B5/Conv 2 &  B5/Conv 1  & B5/Conv 2  &  $1\times1\times1\times7$ & (2,1) \\			
			&    B5/Conv 3 &  B5/Conv 2  & B5/Conv 3  &  $1\times1\times1\times15$ & (1,0) \\	
			&    B5/Conv 4 &  B5/Conv 3  & B5/out   &  $1\times1\times1\times15$ & (1,0) \\
			\hline\hline

			\multirow{5}{*}{Feature Fusion}  & \multirow{5}{*}{Fusion $1$} &  B1/out   & \multirow{5}{*}{Latent}  & \centering \multirow{5}{*}{-} & \multirow{5}{*}{-} \\
			&   &  B2/out &    &    &   \\
			&   &  B3/out &    &    &   \\
			&   &  B4/out &    &    &   \\
			&   &  B5/out &    &    &   \\	
			\hline\hline

			\multirow{1}{*}{Self-expressiveness}  & \multirow{1}{*}{ $\Theta_s$} &  Latent    & L-recon  & \centering \multirow{1}{*}{$4665600$ Parameters} & \multirow{1}{*}{-} \\
			\hline\hline
			\multirow{5}{*}{Multimodal}  & \multirow{5}{*}{Decoder} &  \multirow{5}{*}{L-recon}    & Recon 1  & \centering \multirow{5}{*}{Details in} & \multirow{5}{*}{} \\
			\multirow{5}{*}{Decoder} &\multirow{5}{*}{layers}&  & Recon 2  & \multirow{5}{*}{Table~\ref{tbl:decoder_arl}}& \\
			&  &  & Recon 3  &  & \\
			&  &  & Recon 4  &  & \\
			&  &  & Recon 5  &  & \\	
			\hline	
		\end{tabular} 
	}
\end{table}
\begin{table}[htp!]
	\centering
	\caption{Intermediate spatial fusion Networks in the ARL experiments.}
	\label{table:tbl3r}
	\resizebox{\linewidth}{!}{
		\begin{tabular}{|c|c|c|c|l|  p{0.7cm}|}
			\cline{2-6}
			\multicolumn{1}{c|}{ } & \multirow{3}{*}{Layer} & \multirow{3}{*}{Input} & \multirow{3}{*}{output} & \multirow{3}{*}{Kernel} & (stride, pad)  \\
			\cline{2-6}\hline	
			\multirow{5}{*}{Layer 1}  & B1/Conv  1 &  Image $1$  & B1/Conv 1  &  $1\times3\times3\times5$ & (2,1) \\
			& B2/Conv  1 &  Image $2$  & B2/Conv 1  &  $1\times3\times3\times5$ & (2,1) \\
			& B3/Conv  1 &  Image $3$  & B3/Conv 1  &  $1\times3\times3\times5$ & (2,1) \\
			& B4/Conv  1 &  Image $4$  & B4/Conv 1  &  $1\times3\times3\times5$ & (2,1) \\
			& B5/Conv  1 &  Image $5$  & B5/Conv 1  &  $1\times3\times3\times5$ & (2,1) \\
			
			\hline\hline	
			
			\multirow{3}{*}{Feature Fusion}  & \multirow{3}{*}{B345/Fusion} & B3/Conv 1   & \multirow{3}{*}{B345/Fusion}  & \centering \multirow{3}{*}{-} & \multirow{3}{*}{-} \\
			&   &  B4/Conv 1 &    &    &   \\
			&   &  B5/Conv 1 &    &    &   \\
			
			\hline\hline	
			
			\multirow{3}{*}{Layer 2}&    B1/Conv 2 &  B1/Conv 1  & B1/Conv 2  &  $1\times1\times1\times7$ & (2,1) \\
			&    B2/Conv 2 &  B2/Conv 1  & B2/Conv 2  &  $1\times1\times1\times7$ & (2,1) \\
			&    B345/Conv 2 &  B345/Fusion  & B345/Conv 2  &  $1\times1\times1\times7$ & (2,1) \\	
			\hline\hline
			\multirow{3}{*}{Feature Fusion}  & \multirow{3}{*}{B2345/Fusion} & B345/Conv 2   & \multirow{3}{*}{B2345/Fusion}  & \centering \multirow{3}{*}{-} & \multirow{3}{*}{-} \\
			&   &  B2/Conv 2 &    &    &   \\
			\hline\hline
			
			\multirow{2}{*}{Layer 3}&    B1/Conv 3 &  B1/Conv 2  & B1/Conv 3  &  $1\times1\times1\times15$ & (1,0) \\	
			&    B2345/Conv 3 &  B2345/Fusion  & B2345/Conv 3  &  $1\times1\times1\times15$ & (1,0) \\	
			\hline\hline
			\multirow{2}{*}{Feature Fusion}  & \multirow{2}{*}{Ball/Fusion} & B1/Conv 3   & \multirow{2}{*}{Ball/Fusion}  & \centering \multirow{3}{*}{-} & \multirow{2}{*}{-} \\
			&   &  B2345/Conv 3 &    &    &   \\
			\hline\hline
			\multirow{1}{*}{Layer 4}&    Ball/Conv 4 &  Ball/Conv 3  & Latent  &  $1\times1\times1\times15$ & (1,0) \\
			\hline\hline

			\multirow{1}{*}{Self-expressiveness}  & \multirow{1}{*}{ $\Theta_s$} &  Latent    & L-recon  & \centering \multirow{1}{*}{$4665600$ Parameters} & \multirow{1}{*}{-} \\
			\hline\hline
			\multirow{5}{*}{Multimodal}  & \multirow{5}{*}{Decoder} &  \multirow{5}{*}{L-recon}    & Recon 1  & \centering \multirow{5}{*}{Details in} & \multirow{5}{*}{} \\
			\multirow{5}{*}{Decoder} &\multirow{5}{*}{layers}&  & Recon 2  & \multirow{5}{*}{Table~\ref{tbl:decoder_arl}}& \\
			&  &  & Recon 3  &  & \\
			&  &  & Recon 4  &  & \\
			&  &  & Recon 5  &  & \\

			\hline
			
		\end{tabular} 
	}
\end{table}

\begin{table}[htp!]
	\centering
	\caption{Affinity fusion networks in the ARL experiments.}
	\label{table:tbl3r}
	\resizebox{\linewidth}{!}{
		\begin{tabular}{|c|c|c|c|l|  p{0.7cm}|}
			\cline{2-6}
			\multicolumn{1}{c|}{ } & \multirow{3}{*}{Layer} & \multirow{3}{*}{Input} & \multirow{3}{*}{output} & \multirow{3}{*}{Kernel} & (stride, pad)  \\
			\cline{2-6}\hline	
			\multirow{4}{*}{Encoder 1}  & B1/Conv  1 &  Image $1$  & B1/Conv 1  &  $1\times3\times3\times5$ & (2,1) \\
			&    B1/Conv 2 &  B1/Conv 1  & B1/Conv 2  &  $1\times1\times1\times7$ & (2,1) \\			
			&    B1/Conv 3 &  B1/Conv 2  & B1/Conv 3  &  $1\times1\times1\times15$ & (1,0) \\	
			&    B1/Conv 4 &  B1/Conv 3  & Latent 1  &  $1\times1\times1\times15$ & (1,0) \\
			\hline\hline
			\multirow{4}{*}{Encoder 2}  & B2/Conv 1 &  Image $2$  & B2/Conv 1  &  $1\times3\times3\times5$ & (2,1) \\
			&    B2/Conv 2 &  B2/Conv 1  & B2/Conv 2  &  $1\times1\times1\times7$  & (2,1) \\			
			&    B2/Conv 3 &  B2/Conv 2  & B2/Conv 3  &  $1\times1\times1\times15$ & (1,0) \\	
			&    B2/Conv 4 &  B2/Conv 3  & Latent 2  &  $1\times1\times1\times15$ & (1,0) \\
			\hline\hline		
			
			\multirow{4}{*}{Encoder 3}  & B3/Conv 1 &  Image $3$  & B3/Conv 1  &  $1\times3\times3\times5$ & (2,1) \\
			&    B3/Conv 2 &  B3/Conv 1  & B3/Conv 2  &  $1\times1\times1\times7$  & (2,1) \\			
			&    B3/Conv 3 &  B3/Conv 2  & B3/Conv 3  &  $1\times1\times1\times15$ & (1,0) \\	
			&    B3/Conv 4 &  B3/Conv 3  & Latent 3   &  $1\times1\times1\times15$ & (1,0) \\
			\hline\hline	
			\multirow{4}{*}{Encoder 4}  & B4/Conv 1 &  Image $4$  & B4/Conv 1  &  $1\times3\times3\times5$ & (2,1) \\
			&    B4/Conv 2 &  B4/Conv 1  & B4/Conv 2  &  $1\times1\times1\times7$  & (2,1) \\			
			&    B4/Conv 3 &  B4/Conv 2  & B4/Conv 3  &  $1\times1\times1\times15$ & (1,0) \\	
			&    B4/Conv 4 &  B4/Conv 3  & Latent 4   &  $1\times1\times1\times15$ & (1,0) \\
			\hline\hline		
			\multirow{4}{*}{Encoder 5}  & B5/Conv 1 &  Image $5$  & B5/Conv 1  &  $1\times3\times3\times5$ & (2,1) \\
			&    B5/Conv 2 &  B5/Conv 1  & B5/Conv 2  &  $1\times1\times1\times7$ & (2,1) \\			
			&    B5/Conv 3 &  B5/Conv 2  & B5/Conv 3  &  $1\times1\times1\times15$ & (1,0) \\	
			&    B5/Conv 4 &  B5/Conv 3  & Latent 5  &  $1\times1\times1\times15$ & (1,0) \\
			\hline\hline

			\multirow{5}{*}{Self-expressiveness}  & \multirow{5}{*}{Common $\Theta_s$} &  Latent 1   & L-recon 1  & \centering \multirow{5}{*}{$4665600$ Parameters} & \multirow{5}{*}{-} \\
			\multirow{5}{*}{layer}&   &  Latent 2 &  L-recon 2  &    &   \\
			&   &  Latent 3 &  L-recon 3  &    &   \\
			&   &  Latent 4 &  L-recon 4  &    &   \\
			&   &  Latent 5 &  L-recon 5  &    &   \\	
			\hline\hline	
			\multirow{3}{*}{Decoder 1}  & D1/deconv 1 &  L-recon $1$  & D1/deconv 1  & $1\times1\times1\times15$  & (1,0) \\
			&    D1/deconv 2 &  D1/deconv 1  & D1/deconv 2  &  $1\times1\times1\times7$ & (2,1) \\			
			&    D1/deconv 3 &  D1/deconv 2  & Recon 1  &  $1\times3\times3\times5$ & (2,1) \\	
			\hline\hline
			\multirow{3}{*}{Decoder 2}  & D2/deconv 2 &  L-recon $2$  & D2/deconv 1  & $1\times1\times1\times15$  & (1,0) \\
			&    D2/deconv 2 &  D2/deconv 1  & D2/deconv 2  &  $1\times1\times1\times7$ & (2,1) \\			
			&    D2/deconv 3 &  D2/deconv 2  & Recon 2  &  $1\times3\times3\times5$ & (2,1) \\	
			\hline\hline
			\multirow{3}{*}{Decoder 3}  & D3/deconv 2 &  L-recon $3$  & D3/deconv 1  & $1\times1\times1\times15$  & (1,0) \\
			&    D3/deconv 2 &  D3/deconv 1  & D3/deconv 2  &  $1\times1\times1\times7$ & (2,1) \\			
			&    D3/deconv 3 &  D3/deconv 2  & Recon 3  &  $1\times3\times3\times5$ & (2,1) \\	
			\hline\hline
			\multirow{3}{*}{Decoder 4}  & D4/deconv 2 &  L-recon $4$  & D4/deconv 1  & $1\times1\times1\times15$  & (1,0) \\
			&    D4/deconv 2 &  D4/deconv 1  & D4/deconv 2  &  $1\times1\times1\times7$ & (2,1) \\			
			&    D4/deconv 3 &  D4/deconv 2  & Recon 4 &  $1\times3\times3\times5$ & (2,1) \\	
			\hline\hline
			\multirow{3}{*}{Decoder 5}  & D5/deconv 2 &  L-recon $5$  & D5/deconv 1  & $1\times1\times1\times15$  & (1,0) \\
			&    D5/deconv 2 &  D5/deconv 1  & D5/deconv 2  &  $1\times1\times1\times7$ & (2,1) \\			
			&    D5/deconv 3 &  D5/deconv 2  & Recon 5 &  $1\times3\times3\times5$ & (2,1) \\	
			\hline
			
		\end{tabular} 
	}
\end{table}

\begin{table}[htp!]
	\centering
	\caption{Multimodal decoders in the ARL experiments.}
	\label{tbl:decoder_arl}
	\resizebox{\linewidth}{!}{
		\begin{tabular}{|c|c|c|c|l|  p{0.7cm}|}
			\cline{2-6}
			\multicolumn{1}{c|}{ } & \multirow{3}{*}{Layer} & \multirow{3}{*}{Input} & \multirow{3}{*}{output} & \multirow{3}{*}{Kernel} & (stride, pad)  \\
			\cline{2-6}\hline	
			\multirow{3}{*}{Decoder 1}  & D1/deconv 1 &  L-recon   & D1/deconv 1  & $1\times1\times1\times15$  & (1,0) \\
			&    D1/deconv 2 &  D1/deconv 1  & D1/deconv 2  &  $1\times1\times1\times7$ & (2,1) \\			
			&    D1/deconv 3 &  D1/deconv 2  & Recon 1  &  $1\times3\times3\times5$ & (2,1) \\	
			\hline\hline
			\multirow{3}{*}{Decoder 2}  & D2/deconv 2 &  L-recon   & D2/deconv 1  & $1\times1\times1\times15$  & (1,0) \\
			&    D2/deconv 2 &  D2/deconv 1  & D2/deconv 2  &  $1\times1\times1\times7$ & (2,1) \\			
			&    D2/deconv 3 &  D2/deconv 2  & Recon 2  &  $1\times3\times3\times5$ & (2,1) \\	
			\hline\hline
			\multirow{3}{*}{Decoder 3}  & D3/deconv 2 &  L-recon   & D3/deconv 1  & $1\times1\times1\times15$  & (1,0) \\
			&    D3/deconv 2 &  D3/deconv 1  & D3/deconv 2  &  $1\times1\times1\times7$ & (2,1) \\			
			&    D3/deconv 3 &  D3/deconv 2  & Recon 3  &  $1\times3\times3\times5$ & (2,1) \\	
			\hline\hline
			\multirow{3}{*}{Decoder 4}  & D4/deconv 2 &  L-recon  & D4/deconv 1  & $1\times1\times1\times15$  & (1,0) \\
			&    D4/deconv 2 &  D4/deconv 1  & D4/deconv 2  &  $1\times1\times1\times7$ & (2,1) \\			
			&    D4/deconv 3 &  D4/deconv 2  & Recon 4 &  $1\times3\times3\times5$ & (2,1) \\	
			\hline\hline
			\multirow{3}{*}{Decoder 5}  & D5/deconv 2 &  L-recon  & D5/deconv 1  & $1\times1\times1\times15$  & (1,0) \\
			&    D5/deconv 2 &  D5/deconv 1  & D5/deconv 2  &  $1\times1\times1\times7$ & (2,1) \\			
			&    D5/deconv 3 &  D5/deconv 2  & Recon 5 &  $1\times3\times3\times5$ & (2,1) \\	
			\hline
			
		\end{tabular} 
	}
\end{table}


\pagebreak

\subsection{Different networks corresponding to Extended Yale-B experiments}

\begin{table}[htp!]
	\centering
	\caption{Early-fusion networks in the Extended Yale-B experiments.}
	\label{table:tbl3r}
	\resizebox{\linewidth}{!}{
		\begin{tabular}{|c | c | c|c | l | p{0.7cm}|}
			\cline{2-6}
			\multicolumn{1}{c|}{ } & \multirow{3}{*}{Layer} & \multirow{3}{*}{Input} & \multirow{3}{*}{output} & \multirow{3}{*}{Kernel} & (stride, pad)  \\
			\cline{2-6}\hline			
			\multirow{5}{*}{Feature Fusion}  & \multirow{5}{*}{Fusion $1$} &  Image 1  & \multirow{5}{*}{Fusion $1$}  & \centering \multirow{5}{*}{-} & \multirow{5}{*}{-} \\
			&   & Image 2 &    &    &   \\
			&   & Image 3 &    &    &   \\
			&   & Image 4 &    &    &   \\
			&   & Image 5 &    &    &   \\	
			\hline\hline
			\multirow{4}{*}{Convolutional layers}  & Conv 1 &  Fusion $1$  & Conv 1  &  $1\times5\times5\times10$ & (2,1) \\
			&    Conv 2 &  Conv 1  & Conv 2  &  $1\times3\times3\times20$ & (2,1) \\			
			&    Conv 3 &  Conv 2  & Conv 3  &  $1\times3\times3\times30$ & (1,0) \\	
			&    Conv 4 &  Conv 3  & Latent  &  $1\times3\times3\times30$ & (1,0) \\
			\hline\hline
			
			\multirow{1}{*}{Self-expressiveness}  & \multirow{1}{*}{ $\Theta_s$} &  Latent    & L-recon  & \centering \multirow{1}{*}{$5914624$ Parameters} & \multirow{1}{*}{-} \\
			\hline\hline
			\multirow{5}{*}{Multimodal}  & \multirow{5}{*}{Decoder} &  \multirow{5}{*}{L-recon}    & Recon 1  & \centering \multirow{5}{*}{Details in} & \multirow{5}{*}{} \\
			\multirow{5}{*}{Decoder} &\multirow{5}{*}{layers}&  & Recon 2  & \multirow{5}{*}{Table~\ref{tbl:decoder_yaleb}}& \\
			&  &  & Recon 3  &  & \\
			&  &  & Recon 4  &  & \\
			&  &  & Recon 5  &  & \\	
			\hline	
		\end{tabular} 
	}
\end{table}


\begin{table}[htp!]
	\centering
	\caption{Late-fusion networks in the Extended Yale-B experiments.}
	\label{table:tbl3r}
	\resizebox{\linewidth}{!}{
		\begin{tabular}{|c | c | c|c | c |  p{0.7cm}|}
			\cline{2-6}
			\multicolumn{1}{c|}{ } & \multirow{3}{*}{Layer} & \multirow{3}{*}{Input} & \multirow{3}{*}{output} & \multirow{3}{*}{Kernel} & (stride, pad)  \\
			\cline{2-6}\hline	
			\multirow{4}{*}{Branch 1}  & B1/Conv  1 &  Image $1$  & B1/Conv 1  &  $1\times5\times5\times10$ & (2,1) \\
			&    B1/Conv 2 &  B1/Conv 1  & B1/Conv 2  &  $1\times3\times3\times20$ & (2,1) \\			
			&    B1/Conv 3 &  B1/Conv 2  & B1/Conv 3  &  $1\times3\times3\times30$ & (1,0) \\	
			&    B1/Conv 4 &  B1/Conv 3  & B1/out  &  $1\times3\times3\times30$ & (1,0) \\
			\hline\hline
			\multirow{4}{*}{Branch 2}  & B2/Conv 1 &  Image $2$  & B2/Conv 1  &  $1\times5\times5\times10$ & (2,1) \\
			&    B2/Conv 2 &  B2/Conv 1  & B2/Conv 2  &  $1\times3\times3\times20$ & (2,1) \\			
			&    B2/Conv 3 &  B2/Conv 2  & B2/Conv 3  &  $1\times3\times3\times30$ & (1,0) \\	
			&    B2/Conv 4 &  B2/Conv 3  & B2/out  &  $1\times3\times3\times30$ & (1,0) \\
			\hline\hline		
			
			\multirow{4}{*}{Branch 3}  & B3/Conv 1 &  Image $3$  & B3/Conv 1  &  $1\times5\times5\times10$ & (2,1) \\
			&    B3/Conv 2 &  B3/Conv 1  & B3/Conv 2  &  $1\times3\times3\times20$ & (2,1) \\			
			&    B3/Conv 3 &  B3/Conv 2  & B3/Conv 3  &  $1\times3\times3\times30$ & (1,0) \\	
			&    B3/Conv 4 &  B3/Conv 3  & B3/out   &  $1\times3\times3\times30$ & (1,0) \\
			\hline\hline	
			\multirow{4}{*}{Branch 4}  & B4/Conv 1 &  Image $4$  & B4/Conv 1  &  $1\times5\times5\times10$ & (2,1) \\
			&    B4/Conv 2 &  B4/Conv 1  & B4/Conv 2  &  $1\times3\times3\times20$ & (2,1) \\			
			&    B4/Conv 3 &  B4/Conv 2  & B4/Conv 3  &  $1\times3\times3\times30$ & (1,0) \\	
			&    B4/Conv 4 &  B4/Conv 3  & B4/out   &  $1\times3\times3\times30$ & (1,0) \\
			\hline\hline		
			\multirow{4}{*}{Branch 5}  & B5/Conv 1 &  Image $5$  & B5/Conv 1  &  $1\times5\times5\times10$ & (2,1) \\
			&    B5/Conv 2 &  B5/Conv 1  & B5/Conv 2  &  $1\times3\times3\times20$ & (2,1) \\			
			&    B5/Conv 3 &  B5/Conv 2  & B5/Conv 3  &  $1\times3\times3\times30$ & (1,0) \\	
			&    B5/Conv 4 &  B5/Conv 3  & B5/out   &  $1\times3\times3\times30$ & (1,0) \\
			\hline\hline

			\multirow{5}{*}{Feature Fusion}  & \multirow{5}{*}{Fusion $1$} &  B1/out   & \multirow{5}{*}{Latent}  & \centering \multirow{5}{*}{-} & \multirow{5}{*}{-} \\
			&   &  B2/out &    &    &   \\
			&   &  B3/out &    &    &   \\
			&   &  B4/out &    &    &   \\
			&   &  B5/out &    &    &   \\	
			\hline\hline
			
			\multirow{1}{*}{Self-expressiveness}  & \multirow{1}{*}{ $\Theta_s$} &  Latent    & L-recon  & \centering \multirow{1}{*}{$5914624$ Parameters} & \multirow{1}{*}{-} \\
			\hline\hline
			\multirow{5}{*}{Multimodal}  & \multirow{5}{*}{Decoder} &  \multirow{5}{*}{L-recon}    & Recon 1  & \centering \multirow{5}{*}{Details in} & \multirow{5}{*}{} \\
			\multirow{5}{*}{Decoder} &\multirow{5}{*}{layers}&  & Recon 2  & \multirow{5}{*}{Table~\ref{tbl:decoder_yaleb}}& \\
			&  &  & Recon 3  &  & \\
			&  &  & Recon 4  &  & \\
			&  &  & Recon 5  &  & \\	
			\hline	
		\end{tabular} 
	}
\end{table}


\begin{table}[htp!]
	\centering
	\caption{Affinity fusion networks in the Extended Yale-B experiments.}
	\label{table:tbl3r}
	\resizebox{\linewidth}{!}{
		\begin{tabular}{|c|c|c|c|l|  p{0.7cm}|}
			\cline{2-6}
			\multicolumn{1}{c|}{ } & \multirow{3}{*}{Layer} & \multirow{3}{*}{Input} & \multirow{3}{*}{output} & \multirow{3}{*}{Kernel} & (stride, pad)  \\
			\cline{2-6}\hline	
			\multirow{4}{*}{Encoder 1}  & B1/Conv  1 &  Image $1$  & B1/Conv 1  &  $1\times5\times5\times10$ & (2,1) \\
			&    B1/Conv 2 &  B1/Conv 1  & B1/Conv 2  &  $1\time3\times3\times20$ & (2,1) \\			
			&    B1/Conv 3 &  B1/Conv 2  & B1/Conv 3  &  $1\times3\times3\times30$ & (1,0) \\	
			&    B1/Conv 4 &  B1/Conv 3  & Latent 1  &  $1\times3\times3\times30$ & (1,0) \\
			\hline\hline
			\multirow{4}{*}{Encoder 2}  & B2/Conv 1 &  Image $2$  & B2/Conv 1  &  $1\times5\times5\times10$ & (2,1) \\
			&    B2/Conv 2 &  B2/Conv 1  & B2/Conv 2  &  $1\times3\times3\times20$  & (2,1) \\			
			&    B2/Conv 3 &  B2/Conv 2  & B2/Conv 3  &  $1\times3\times3\times30$ & (1,0) \\	
			&    B2/Conv 4 &  B2/Conv 3  & Latent 2  &  $1\times3\times3\times30$ & (1,0) \\
			\hline\hline		
			
			\multirow{4}{*}{Encoder 3}  & B3/Conv 1 &  Image $3$  & B3/Conv 1  &  $1\times5\times5\times10$ & (2,1) \\
			&    B3/Conv 2 &  B3/Conv 1  & B3/Conv 2  &  $1\times3\times3\times20$  & (2,1) \\			
			&    B3/Conv 3 &  B3/Conv 2  & B3/Conv 3  &  $1\times3\times3\times30$ & (1,0) \\	
			&    B3/Conv 4 &  B3/Conv 3  & Latent 3   &  $1\times3\times3\times30$ & (1,0) \\
			\hline\hline	
			\multirow{4}{*}{Encoder 4}  & B4/Conv 1 &  Image $4$  & B4/Conv 1  &  $1\times5\times5\times10$ & (2,1) \\
			&    B4/Conv 2 &  B4/Conv 1  & B4/Conv 2  &  $1\times3\times3\times20$  & (2,1) \\			
			&    B4/Conv 3 &  B4/Conv 2  & B4/Conv 3  &  $1\times3\times3\times30$ & (1,0) \\	
			&    B4/Conv 4 &  B4/Conv 3  & Latent 4  &  $1\times3\times3\times30$ & (1,0) \\
			\hline\hline		
			\multirow{4}{*}{Encoder 5}  & B5/Conv 1 &  Image $5$  & B5/Conv 1  &  $1\times5\times5\times10$ & (2,1) \\
			&    B5/Conv 2 &  B5/Conv 1  & B5/Conv 2  &  $1\times3\times3\times20$ & (2,1) \\			
			&    B5/Conv 3 &  B5/Conv 2  & B5/Conv 3  &  $1\times3\times3\times30$ & (1,0) \\	
			&    B5/Conv 4 &  B5/Conv 3  & Latent 5   &  $1\times3\times3\times30$ & (1,0) \\
			\hline\hline

			\multirow{5}{*}{Self-expressiveness}  & \multirow{5}{*}{Common $\Theta_s$} &  Latent 1   & L-recon 1  & \centering \multirow{5}{*}{$5914624$ Parameters} & \multirow{5}{*}{-} \\
			\multirow{5}{*}{layer}&   &  Latent 2 &  L-recon 2  &    &   \\
			&   &  Latent 3 &  L-recon 3  &    &   \\
			&   &  Latent 4 &  L-recon 4  &    &   \\
			&   &  Latent 5 &  L-recon 5  &    &   \\	
			\hline\hline	
			\multirow{3}{*}{Decoder 1}  & D1/deconv 1 &  L-recon $1$  & D1/deconv 1  & $1\times3\times3\times30$  & (1,0) \\
			&    D1/deconv 2 &  D1/deconv 1  & D1/deconv 2  &  $1\times3\times3\times20$ & (2,1) \\			
			&    D1/deconv 3 &  D1/deconv 2  & Recon 1  &  $1\times5\times5\times10$ & (2,1) \\	
			\hline\hline
			\multirow{3}{*}{Decoder 2}  & D2/deconv 2 &  L-recon $2$  & D2/deconv 1  & $1\times3\times3\times30$  & (1,0) \\
			&    D2/deconv 2 &  D2/deconv 1  & D2/deconv 2  &  $1\times3\times3\times20$ & (2,1) \\			
			&    D2/deconv 3 &  D2/deconv 2  & Recon 2  &  $1\times5\times5\times10$ & (2,1) \\	
			\hline\hline
			\multirow{3}{*}{Decoder 3}  & D3/deconv 2 &  L-recon $3$  & D3/deconv 1  & $1\times3\times3\times30$  & (1,0) \\
			&    D3/deconv 2 &  D3/deconv 1  & D3/deconv 2  &  $1\times3\times3\times20$ & (2,1) \\			
			&    D3/deconv 3 &  D3/deconv 2  & Recon 3  &  $1\times5\times5\times10$ & (2,1) \\	
			\hline\hline
			\multirow{3}{*}{Decoder 4}  & D4/deconv 2 &  L-recon $4$  & D4/deconv 1  & $1\times3\times3\times30$  & (1,0) \\
			&    D4/deconv 2 &  D4/deconv 1  & D4/deconv 2  &  $1\times3\times3\times20$ & (2,1) \\			
			&    D4/deconv 3 &  D4/deconv 2  & Recon 4 &  $1\times5\times5\times10$ & (2,1) \\	
			\hline\hline
			\multirow{3}{*}{Decoder 5}  & D5/deconv 2 &  L-recon $5$  & D5/deconv 1  & $1\times3\times3\times30$  & (1,0) \\
			&    D5/deconv 2 &  D5/deconv 1  & D5/deconv 2  &  $1\times3\times3\times20$ & (2,1) \\			
			&    D5/deconv 3 &  D5/deconv 2  & Recon 5 &  $1\times5\times5\times10$ & (2,1) \\	
			\hline
			
		\end{tabular} 
	}
\end{table}

\begin{table}[htp!]
	\centering
	\caption{Intermediate spatial fusion Networks in the Extended Yale-B experiments.}
	\label{table:tbl3r}
	\resizebox{\linewidth}{!}{
		\begin{tabular}{|c|c|c|c|l|  p{0.7cm}|}
			\cline{2-6}
			\multicolumn{1}{c|}{ } & \multirow{3}{*}{Layer} & \multirow{3}{*}{Input} & \multirow{3}{*}{output} & \multirow{3}{*}{Kernel} & (stride, pad)  \\
			\cline{2-6}\hline	
			\multirow{5}{*}{Layer 1}  & B1/Conv  1 &  Image $1$  & B1/Conv 1  &  $1\times5\times5\times10$ & (2,1) \\
			& B2/Conv  1 &  Image $2$  & B2/Conv 1  &  $1\times5\times5\times10$ & (2,1) \\
			& B3/Conv  1 &  Image $3$  & B3/Conv 1  &  $1\times5\times5\times10$ & (2,1) \\
			& B4/Conv  1 &  Image $4$  & B4/Conv 1  &  $1\times5\times5\times10$ & (2,1) \\
			& B5/Conv  1 &  Image $5$  & B5/Conv 1  &  $1\times5\times5\times10$ & (2,1) \\
			
			\hline\hline	
			
			\multirow{4}{*}{Feature Fusion}  & \multirow{3}{*}{B23/Fusion} & B2/Conv 1   & \multirow{3}{*}{B23/Fusion}  & \centering \multirow{3}{*}{-} & \multirow{3}{*}{-} \\
			&   &  B3/Conv 1 &    &    &   \\
			\cline{2-6}
			& \multirow{3}{*}{B45/Fusion} & B4/Conv 1   & \multirow{3}{*}{B45/Fusion}  & \centering \multirow{3}{*}{-} & \multirow{3}{*}{-} \\
			&   &  B5/Conv 1 &    &    &   \\
			
			\hline\hline	
			
			\multirow{3}{*}{Layer 2}&    B1/Conv 2 &  B1/Conv 1  & B1/Conv 2  &  $1\times3\times3\times20$ & (2,1) \\
			&    B23/Conv 2 &  B23/Fusion  & B23/Conv 2  &  $1\times3\times3\times20$ & (2,1) \\	
			&    B45/Conv 2 &  B45/Fusion  & B45/Conv 2  &  $1\times3\times3\times20$ & (2,1) \\	
			\hline\hline
			\multirow{3}{*}{Feature Fusion}  & \multirow{3}{*}{B2345/Fusion} & B23/Conv 2   & \multirow{3}{*}{B2345/Fusion}  & \centering \multirow{3}{*}{-} & \multirow{3}{*}{-} \\
			&   &  B45/Conv 2 &    &    &   \\
			\hline\hline
			
			\multirow{2}{*}{Layer 3}&    B1/Conv 3 &  B1/Conv 2  & B1/Conv 3  &  $1\times3\times3\times30$ & (1,0) \\	
			&    B2345/Conv 3 &  B2345/Fusion  & B2345/Conv 3  &  $1\times3\times3\times30$ & (1,0) \\	
			\hline\hline
			\multirow{2}{*}{Feature Fusion}  & \multirow{3}{*}{Ball/Fusion} & B1/Conv 3   & \multirow{3}{*}{Ball/Fusion}  & \centering \multirow{3}{*}{-} & \multirow{3}{*}{-} \\
			&   &  B2345/Conv 3 &    &    &   \\
			\hline\hline
			\multirow{1}{*}{Layer 4}&    Ball/Conv 4 &  Ball/Conv 3  & Latent  &  $1\times3\times3\times30$ & (1,0) \\
			\hline\hline

			\multirow{1}{*}{Self-expressiveness}  & \multirow{1}{*}{ $\Theta_s$} &  Latent    & L-recon  & \centering \multirow{1}{*}{$5914624$ Parameters} & \multirow{1}{*}{-} \\
			\hline\hline
			\multirow{5}{*}{Multimodal}  & \multirow{5}{*}{Decoder} &  \multirow{5}{*}{L-recon}    & Recon 1  & \centering \multirow{5}{*}{Details in} & \multirow{5}{*}{} \\
			\multirow{5}{*}{Decoder} &\multirow{5}{*}{layers}&  & Recon 2  & \multirow{5}{*}{Table~\ref{tbl:decoder_yaleb}}& \\
			&  &  & Recon 3  &  & \\
			&  &  & Recon 4  &  & \\
			&  &  & Recon 5  &  & \\

			\hline
			
		\end{tabular} 
	}
\end{table}

\begin{table}[htp!]
	\centering
	\caption{Multimodal decoder details in the Extended Yale-B experiments.}
	\label{tbl:decoder_yaleb}
	\resizebox{\linewidth}{!}{
		\begin{tabular}{|c|c|c|c|l|  p{0.7cm}|}
			\cline{2-6}
			\multicolumn{1}{c|}{ } & \multirow{3}{*}{Layer} & \multirow{3}{*}{Input} & \multirow{3}{*}{output} & \multirow{3}{*}{Kernel} & (stride, pad)  \\
			\cline{2-6}\hline	
			
			\multirow{3}{*}{Decoder 1}  & D1/deconv 1 &  L-recon & D1/deconv 1  & $1\times3\times3\times30$  & (1,0) \\
			&    D1/deconv 2 &  D1/deconv 1  & D1/deconv 2  &  $1\times3\times3\times20$ & (2,1) \\			
			&    D1/deconv 3 &  D1/deconv 2  & Recon 1  &  $1\times5\times5\times10$ & (2,1) \\	
			\hline\hline
			\multirow{3}{*}{Decoder 2}  & D2/deconv 2 &  L-recon  & D2/deconv 1  & $1\times3\times3\times30$  & (1,0) \\
			&    D2/deconv 2 &  D2/deconv 1  & D2/deconv 2  &  $1\times3\times3\times20$ & (2,1) \\			
			&    D2/deconv 3 &  D2/deconv 2  & Recon 2  &  $1\times5\times5\times10$ & (2,1) \\	
			\hline\hline
			\multirow{3}{*}{Decoder 3}  & D3/deconv 2 &  L-recon  & D3/deconv 1  & $1\times3\times3\times30$  & (1,0) \\
			&    D3/deconv 2 &  D3/deconv 1  & D3/deconv 2  &  $1\times3\times3\times20$ & (2,1) \\			
			&    D3/deconv 3 &  D3/deconv 2  & Recon 3  &  $1\times5\times5\times10$ & (2,1) \\	
			\hline\hline
			\multirow{3}{*}{Decoder 4}  & D4/deconv 2 &  L-recon  & D4/deconv 1  & $1\times3\times3\times30$  & (1,0) \\
			&    D4/deconv 2 &  D4/deconv 1  & D4/deconv 2  &  $1\times3\times3\times20$ & (2,1) \\			
			&    D4/deconv 3 &  D4/deconv 2  & Recon 4 &  $1\times5\times5\times10$ & (2,1) \\	
			\hline\hline
			\multirow{3}{*}{Decoder 5}  & D5/deconv 2 &  L-recon  & D5/deconv 1  & $1\times3\times3\times30$  & (1,0) \\
			&    D5/deconv 2 &  D5/deconv 1  & D5/deconv 2  &  $1\times3\times3\times20$ & (2,1) \\			
			&    D5/deconv 3 &  D5/deconv 2  & Recon 5 &  $1\times5\times5\times10$ & (2,1) \\	
			\hline
			
		\end{tabular} 
	}
\end{table}

\bibliographystyle{IEEEtran}
\bibliography{DMSC}
\begin{IEEEbiography}[{\includegraphics[width=1in,height=1.25in,clip,keepaspectratio]{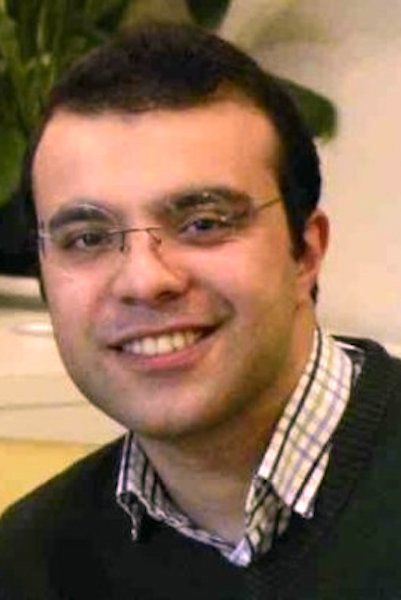}}]{Mahdi Abavisani}
[S'11] received his M.S. degrees in Electrical and Computer Engineering (ECE) from Iran University of Science and Technology, Tehran, Iran in 2014, and Rutgers University, NJ, USA, in 2018. He is currently a Ph.D. candidate in  Electrical and Computer Engineering at Rutgers University. During his Ph.D. he has spent time at Microsoft Research $\&$ AI, and Tesla's Autopilot team in designing deep neural networks for various applications. His research interests include signal and image processing, computer vision, machine learning  and  deep learning.
\end{IEEEbiography}

\begin{IEEEbiography}[{\includegraphics[width=1in,height=1.25in,clip,keepaspectratio]{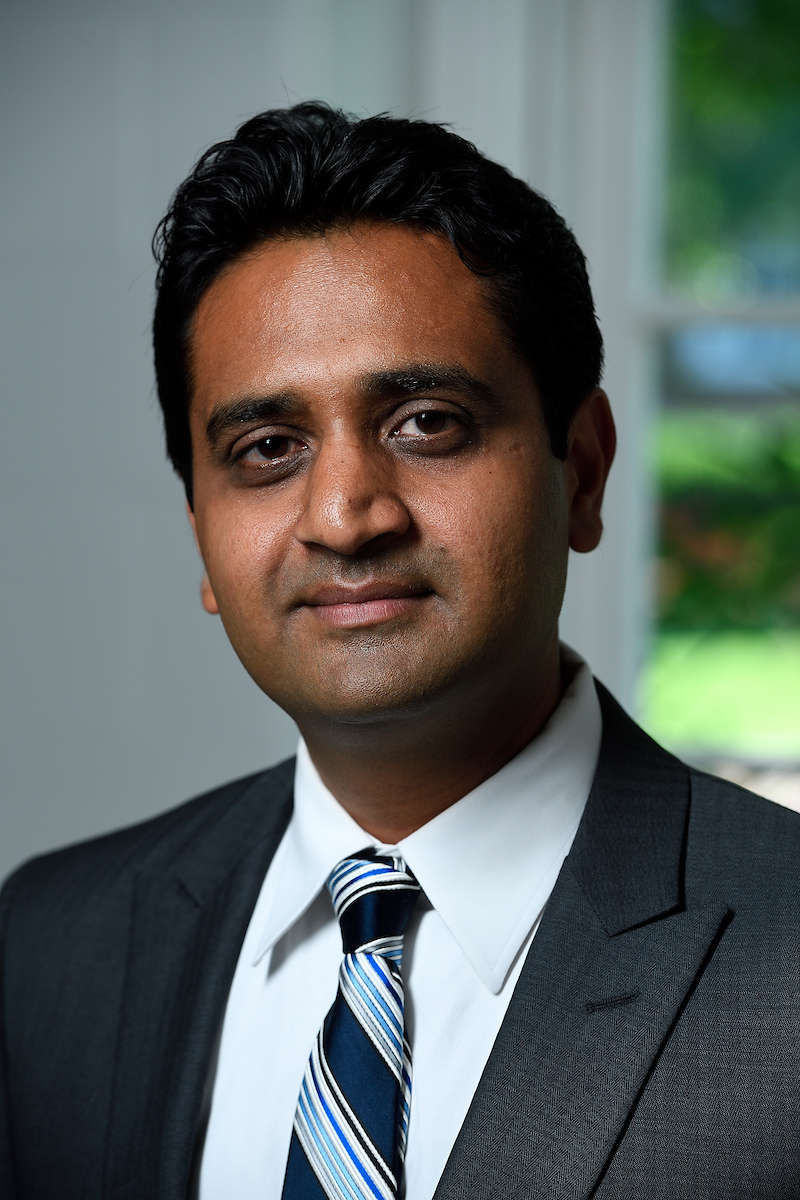}}]{Vishal M. Patel}
Vishal M. Patel [SM'15] is an Assistant Professor in the Department of Electrical and Computer Engineering (ECE) at Johns Hopkins University.  Prior to joining Hopkins, he was an A. Walter Tyson Assistant Professor in the Department of ECE at Rutgers University and a member of the research faculty at the University of Maryland Institute for Advanced Computer Studies (UMIACS). He completed his Ph.D. in Electrical Engineering from the University of Maryland, College Park, MD, in 2010. His current research interests include signal processing, computer vision, and pattern recognition with applications in biometrics and imaging. He has received a number of awards including the 2016 ONR Young Investigator Award, the 2016 Jimmy Lin Award for Invention, A. Walter Tyson Assistant Professorship Award, Best Paper Award at IEEE AVSS 2017, Best Paper Award at IEEE BTAS 2015, Honorable Mention Paper Award at IAPR ICB 2018, two Best Student Paper Awards at IAPR ICPR 2018, and Best Poster Awards at BTAS 2015 and 2016. He is an Associate Editor of the IEEE Signal Processing Magazine, IEEE Biometrics Compendium, and serves on the Information Forensics and Security Technical Committee of the IEEE Signal Processing Society. He is a member of Eta Kappa Nu, Pi Mu Epsilon, and Phi Beta Kappa.
\end{IEEEbiography}
\end{document}